\documentclass[final,5p,times,twocolumn]{elsarticle}

\usepackage{lipsum}
\usepackage{graphicx}
\usepackage{booktabs}
\usepackage{makecell}
\usepackage{bm}
\usepackage{amsmath,amssymb,amsfonts,mathtools}
\usepackage{pifont}
\usepackage{caption}
\usepackage{subcaption}
\usepackage{graphics}   
\usepackage{adjustbox}
\usepackage{hyperref}
\usepackage{algorithm}
\usepackage{algpseudocode}
\usepackage{multirow}
\usepackage[usenames,dvipsnames]{xcolor}
\definecolor{dimGreen}{rgb}{0.0, 0.8, 0.0}
\definecolor{dimRed}{rgb}{0.8, 0.0, 0.0}
\usepackage{bbm}
\usepackage{gensymb}

\usepackage[usestackEOL]{stackengine}
\algnewcommand{\Inputs}[1]{%
  \State \textbf{Inputs:}
  \Statex \hspace*{\algorithmicindent}\parbox[t]{.8\linewidth}{\raggedright #1}
}
\algnewcommand{\Initialize}[1]{%
  \State \textbf{Initialize:}
  \Statex \hspace*{\algorithmicindent}\parbox[t]{.8\linewidth}{\raggedright #1}
}
\usepackage{etoolbox}

\biboptions{numbers,sort&compress}

\DeclarePairedDelimiter\autobracket{(}{)}
\newcommand{\br}[1]{\autobracket*{#1}}

\newcommand{\bx}{\bm{x}}
\newcommand{\bth}{\bm{\theta}}
\newcommand{\brth}{\br{\bm{\theta}}}

\newcommand{\D}{\mathcal{D}}
\newcommand{\bz}{{\bm{\zeta}}}

\renewcommand{\P}{\text{P}}

\usepackage{dcolumn}
\newcolumntype{d}[1]{D{.}{.}{#1}}

\def\ie{{\em i.e.},\ }
\def\eg{{\em e.g.},\ }
\long\def\comment#1{}

\journal{Expert Systems with Applications}

\begin{document}

\begin{frontmatter}

\title{RULSurv: A probabilistic survival-based method for early censoring-aware prediction of remaining useful life in ball bearings}

\author{Christian Marius Lillelund}
\ead{cl@ece.au.dk}
\author{Fernando Pannullo}
\ead{202102261@post.au.dk}
\author{Morten Opprud Jakobsen}
\ead{morten@ece.au.dk}
\author{Manuel Morante}
\ead{morante@ece.au.dk}
\author{Christian Fischer Pedersen}
\ead{cfp@ece.au.dk}
\address{Department of Electrical and Computer Engineering, Aarhus University, Finlandsgade 22, 8200 Aarhus N, Denmark}

\begin{abstract}
Predicting the remaining useful life (RUL) of ball bearings is an active area of research, where novel machine learning techniques are continuously being applied to predict degradation trends and anticipate failures before they occur. However, few studies have explicitly addressed the challenge of handling censored data, where information about a specific event (\eg mechanical failure) is incomplete or only partially observed. To address this issue, we introduce a novel and flexible method for early fault detection using Kullback-Leibler (KL) divergence and RUL estimation using survival analysis that naturally supports censored data. We demonstrate our approach in the XJTU-SY dataset using a 5-fold cross-validation strategy across three different operating conditions. When predicting the time to failure for bearings under the highest load (C1, 12.0 kN and 2100 RPM) with 25\% random censoring, our approach achieves a mean absolute error (MAE) of 14.7 minutes (95\% CI = 13.6-15.8) using a linear CoxPH model, and an MAE of 12.6 minutes (95\% CI = 11.8-13.4) using a nonlinear Random Survival Forests model, compared to an MAE of 18.5 minutes (95\% CI = 17.4-19.6) using a linear LASSO model that does not support censoring. Moreover, our approach achieves a mean cumulative relative accuracy (CRA) of 0.7586 over 5 bearings under the highest load, which improves over several state-of-the-art baselines. Our work highlights the importance of considering censored data as part of the model design when building predictive models for early fault detection and RUL estimation.
\end{abstract}


\end{frontmatter}

\section{Introduction}

Rolling bearings are commonly used in various rotating machinery, such as pumps, electric motors, wind turbines, and vehicles, but they are also among the most failure-prone components in these systems. Consequently, engineers and scientists have tried to detect faults early in these components, so that timely maintenance and repairs can be performed. Faults and defects are often induced by contamination wear, poor lubrication, and improper mounting, among others~\citep{CERRADA2018169, kim_prognostics_2017, JIA2018619, cubillo_review_2016}. However, accurately identifying the presence of a fault can be challenging in practical scenarios, particularly when the fault is in its early stages and the signal-to-noise ratio is low.

A defect primarily induces mechanical vibrations that can be detected by an accelerometer or a microphone. Mechanical engineers have traditionally used model-based methods, which are based solely on threshold values of different signals at predetermined fault frequencies. These models can only describe the signal characteristics of a few specific types of faults, but faults in the real world are often more complicated. Thus, researchers have adopted data-driven approaches using machine learning (ML) and deep learning (DL)~\citep{filippetti_recent_2000, awadallah_application_2003, batista_classifier_2013, liu_artificial_2018}, as they offer more precise means of assessing the health of a bearing~\citep{tobon-mejia_data-driven_2012, liewald_perspectives_2022}.

An important metric in bearing diagnostics is the remaining useful life (RUL). The RUL is defined as the time interval between the current inspection time and a future point in time at which the bearing reaches the end of its useful life or fails. The RUL is considered an important indicator of the health of a bearing and can be used to detect mechanical faults before they lead to actual failure~\citep{guo_recurrent_2017, zheng_data-driven_2018, al_masry_remaining_2019, wang_hybrid_2020, wang_remaining_2022, xu_rul_2022, xu_novel_2023}. Accurately estimating the RUL can help engineers plan maintenance in advance.

However, despite numerous advances in model development and increases in predictive performances, one remaining challenge is how to handle censored data, where the event of interest is only partially observed -- \ie for some instances, we only have a lower bound on the actual time of the event. Censored or incomplete observations are a prevalent phenomenon given the rarity of failures or breakdowns in ball bearings~\citep{ferreira_remaining_2022}, but so far, only few studies have attempted to model or address the consequences of censoring~\citep{widodo_machine_2011, widodo_application_2011, hochstein_survival_2013, zhang_time-dependent_2019} and the literature on censored data is limited. We can choose to ignore censored observations and analyze only the data where we know the true outcome, but this would lead to a loss of information and introduce an estimation bias in the model parameters~\citep{stepanova_censoring_2002}. If we want to accurately estimate when a bearing is likely to break down based on historical data and current sensor readings, we should include information from censored data samples in the model. Moreover, since bearing failure is generally a rare event, in the sense that the number of data samples where the bearing operates without problems far exceeds the number of samples with failure indications~\citep{ferreira_remaining_2022}, it is important to make use of the data we have at our disposal, including censored and uncensored observations.

\textbf{Contribution:} In this paper, we extend the work of \cite{lillelund_predicting_2023} and propose a method for RUL prediction consisting of two main components: An interpretable event detection algorithm and a suite of survival models for RUL prediction. The purpose of the event algorithm is to detect significant discrepancies in the frequencies of a bearing, indicating a potential future mechanical failure. We use this algorithm as an oracle to label a ball bearing dataset for subsequent RUL estimation. Specifically, for event detection, we compare the Kullback-Leibler (KL) divergence between the magnitude spectrum of the current signal and the magnitude spectrum of a reference signal. This divergence becomes significant as the bearing starts to deteriorate. Thus, it is a useful indicator of the health of a bearing. For RUL estimation, we propose one of multiple feature-based survival models that (1) naturally supports right-censored data, \ie where the event of interest (bearing failure) is not observed, (2) can give individual bearing predictions, (3) provides survival (event) probability curves that are guaranteed to be monotonically decreasing over time, and (4) predicts the time to failure (\ie the onset of degradation) rather than the end of life. In this context, descriptive features are extracted from the time domain and include information such as skewness, kurtosis, and entropy. For our empirical analyzes, we adopt the XJTU-SY bearing dataset~\citep{wang_hybrid_2020}, which consists of five ball bearings per operating condition (high, medium, low), and we show that our method can reliably detect bearing faults early and also accurately estimate the RUL as the time to failure. In addition, we show that incorporating censored data during training significantly improves the mean absolute error (MAE) compared to methods that do not support censored data, under varying load conditions and censoring levels. Concerning previous work, our approach achieves state-of-the-art performance in terms of cumulative relative accuracy (CRA) on the XJTU-SY dataset under the highest load.

\textbf{Novelty:} Our method has the following novelties: (1) It is based on survival analysis, which can handle the problem of censored observations; in the current literature, there are relatively few RUL prediction methods that support censoring, and most either require timeseries data~\citep{zhang_time-dependent_2019, wang_remaining_2022}, assume proportional hazards (\ie each feature has a time-independent multiplicative effect on the risk)~\citep{wang_remaining_2022, CHEN2020101054}, or do not predict time to failure (\eg minutes, hours), but simply the probability of remaining failure-free at some time in the future~\citep{zhang_time-dependent_2019}. (2) Evaluation of these previous works is often limited, as some only provide aggregated prediction results on a group of bearings~\citep{WANG2023109747, SU2021107531}, or only provide a single prediction error across all bearings~\citep{xu_novel_2023}. Additionally, to the best of our knowledge, no prior work made a direct comparison between censoring-aware methods and methods that do not support censoring for RUL prediction, which we propose. (3) Finally, using survival-based models, we overcome a common limitation in previous works~\citep{xu_novel_2023, zhang_time-dependent_2019, LIN2024102524}, where the predicted RUL does not decrease consistently as a function of time. Our method predicts the RUL as the so-called survival function, which is guaranteed to be monotonically decreasing.

All our experiments are made with Python and are fully reproducible. The source code and data are publicly available at: \url{https://github.com/thecml/rulsurv}

\section{Related work}

Event detection focuses on detecting when a bearing is likely to fail and understanding the relationship between bearing faults and measurable signals. This is usually done by analyzing readings from a bearing sensor, which includes vibration and acoustic noise measurements, and then establishing a threshold value of each input reading. RUL estimation aims to proactively predict the remaining time that a bearing can operate before it reaches the end of its useful life or fails. This is typically done by adopting a data-driven approach that learns patterns in historical data to estimate the RUL of new and unseen bearings. In this section, we provide a brief overview of the latest research in event detection and RUL prediction, the limitations, and motivate our approach.

\subsection{Event detection}

Numerous algorithms have been proposed to detect failures in ball bearings. These include Artificial Neural Networks (ANNs)~\citep{TIAN2015175}, Principal Component Analysis (PCA)~\citep{DONG20133143}, K-Nearest Neighbors (KNN)~\citep{AN201942}, Convolutional Neural Networks (CNNs)~\citep{cheng_convolutional_2021}, Deep Autoencoders~\citep{xu_novel_2023}, Recurrent Neural Networks (RNNs)~\citep{s20185112} and attention-based methods~\citep{SU2021107531}. Although DL approaches have outperformed classical approaches in terms of predictive accuracy, they usually require large amounts of data and inherently lack interpretability: it is practically impossible for humans to trace the precise mapping from input data to prediction in a neural network~\citep{GoodBengCour16, theodoridis2020machine}, and therefore model-specific interpretation methods are needed~\citep{MONTAVON20181}. However, one of the classical approaches that remains relevant is the KL divergence~\citep{kullback1951information}. The KL divergence is a distance-based metric that measures the difference (entropy) between two probability distributions. It can detect anomalous events by comparing the probability density function (PDF) of the current process with a reference PDF. The KL divergence has been widely adopted as an indicator of health degradation~\citep{ZENG20142777, DELPHA2017118, QIN2021108900, WU2023103208}.

\subsection{RUL prediction}

Practitioners and engineers often make RUL predictions by adopting a model-based or data-driven approach: model-based approaches involve developing mathematical or physical models based on historical data to determine trends in the health status of a component~\citep{cheng_convolutional_2021}. On the other hand, a data-driven approach develops a model on historical data and then determines patterns in unseen data~\citep{cheng_convolutional_2021}. This includes ML and DL methods~\citep{ferreira_remaining_2022}. Similarly, despite their need for large amounts of historical data during the training phase, data-driven methods are less complex, more precise, and more applicable in the real world than model-based methods~\citep{tobon-mejia_data-driven_2012, liewald_perspectives_2022}. The availability of large datasets and newer technologies to reliably record sensor readings has also fueled the popularity of data-driven methods, and notable works also rely on a DL architecture for RUL prediction.

\cite{widodo_application_2011} presented a data-driven method to assess machine degradation and directly addressed the problem of censored data by training a Relevance Vector Machine (RVM) using survival analysis. Survival analysis is a form of regression modeling that studies the time to an event, which can be partially observed (that is, censored) and has found important use in applications in multiple domains, such as healthcare informatics~\citep{lillelund_efficient_2024}, econometrics~\citep{stepanova_survival_2002}, and in engineering, including predictive maintenance~\citep{widodo_application_2011, widodo_machine_2011, hochstein_survival_2013, wang_predictive_2017, lillelund_predicting_2023}. The main idea proposed by~\cite{widodo_application_2011} revolves around predicting the survival probability (\ie event probability), indicating when an individual machine component is likely to fail as a function of the measurement points. The method provides an intuitive explanation of a bearing's anticipated failure rate, highlighting distinct declines in bearing health which lead to eventual failure with quantitative measurements. However, the study did not compare the proposed method with others and their results are based only on survival curves derived from the nonparametric Kaplan-Meier (KM) estimator~\citep{kaplan_nonparametric_1958}. Although mathematically robust, nonparametric tests are often difficult to interpret, and the KM estimator does not support features.

\cite{guo_recurrent_2017} proposed a RNN to predict the future health of a bearing. This was achieved by first calculating similarity measures between the current data reading and the data from an initial operation point, scaling this feature from zero to one, and then training the RNN to map this feature value to a corresponding degradation percentage. They trained their proposed method, RNN-HI, on the PRONOSTIA dataset~\citep{nectoux2012pronostia}, and showed significant improvements in RUL prediction accuracy compared to self-organizing maps~\citep{huang_residual_2007}.

\cite{zhang_time-dependent_2019} proposed a time-dependent survival neural network (TSNN), which estimates the risk of failure of a mechanical component by performing multiple classification steps in succession. The output of TSNN is an RUL-specific probability, which is monotonically decreasing. Their approach incorporates temporal dependencies between measurement points, and inherently supports censoring, but the method does not predict the time to failure, rather the probability of remaining failure-free by some time in the future. Additionally, the TSNN model requires a number of observational data points at prediction time. However, observational data require determining features beyond the observed time frame. As a result, time-variant predictions are difficult to make and survival bias may be introduced~\citep{zhou2005survival}.

\cite{CHEN2020101054} proposed a Cox proportional hazards model (CoxPH) that can predict the time between failures in automobiles. They used an auto-encoder architecture to represent the data, a CoxPH model to label the censored observations, and a Long Short-Term Memory (LSTM) network to train a prediction model. However, the proposed CoxPH model is used only to label the censored dataset, and the prediction problem is then framed as a traditional regression problem without paying attention to censoring. It also remains unclear how the CoxPH model was evaluated in terms of predictive accuracy.

\cite{wang_remaining_2022} proposed a CoxPH feature fusion model to predict RUL in rolling bearings, which includes several steps. First, they trained a feature fusion algorithm, which fuses multiple predictors based on sensor readings. Second, they performed a Pearson correlation analysis to obtain a correlation matrix. Third, they trained an LSTM model to produce a set of multidimensional features based on this feature matrix, and finally, the output of the LSTM model is then forwarded to a CoxPH model to obtain the probability of failure for each bearing. The authors provided bearing-specific amplitude readings as predictors and made their predictions as a function of sampling points, showing promise, but did not offer any evaluation of predictive accuracy. The proposed method was evaluated on two bearings only from the Intelligent Maintenance System (IMS) dataset~\citep{lee2007bearing}, but the authors did not provide quantitative performance measurements, and it is unclear when or how they established the event of interest.

\cite{xu_novel_2023} proposed a novel health indicator framework for automatic RUL prediction based on a multi-head attention architecture, combining information from damaged and healthy bearings to improve prediction accuracy. Their method achieves a good prediction error compared to other works, but the estimated RUL is not monotonically decreasing, and the authors make the assumption of a linear correlation between the RUL and a bearing's lifetime. Assuming a linear RUL oversimplifies the bearing degradation process, as the relationship between the RUL and the number of life cycles is often nonlinear~\citep[Ch. 2]{kim_prognostics_2017}. However, bearings may not degrade at a constant speed, as their health can depend on various factors, \eg operating condition, the type of maintenance performed, or other external factors~\citep{5620974}.

\cite{10409180} introduced an improved Sample Convolution and Interaction Network (SCINet) for RUL prediction in ball bearings. The authors first establish a bearing health indicator based on the seuclidean (\ie standardized Euclidean) distance. They then extract descriptive features, \eg the root mean square, skewness, and kurtosis. This data is fused by a multidimensional scale transformation to form a representation of the bearing degradation process. Then, the Isolation Forest algorithm is used to detect the initial degradation point of the bearing, and this information is used to train a RUL prediction model, coined IF-SCINET. The method is evaluated in the XJTU-SY dataset using MAE, \ie the mean absolute difference between the predicted degradation point and the actual degradation point. However, they assume a linear correlation between the RUL and a bearing's lifetime, and the estimated RUL is not monotonically decreasing -- for some intervals, the RUL actually increases as a function of time. Moreover, their method is based on timeseries data, where the prediction target (failure) is fully-observed and never censored.

\cite{LIN2024102524} proposed a hybrid deep learning approach for RUL prediction that utilizes non-full lifecycle (\ie censored) data. The authors combine self-supervised and supervised learning to address data scarcity and noise challenges in RUL prediction tasks. Using a Contrastive Predictive Coding (CPC) encoder and a Transformer decoder, the model leverages non-full lifecycle data for pretraining and finetuning with full lifecycle data. Their experiments show good performance in a separate dataset in terms of the root mean squared error (RMSE). Although censored data are used in the pretraining phase, the prediction task in this case is simply predicting the next feature value $x_{t+1}$ given $x_{t}$, not the time-to-event. In other words, the censored data is not used to predict the probability of future event, but rather to improved the learned feature presentation. Moreover, the authors never evaluate their method with censoring in the testing phase and the proposed RMSE assumes that all samples have observed failure times and the predicted RUL is not monotonically decreasing.

\subsection{Research gap}

Censored data means that the event of interest (failure) is not observed in a sample and we only have a lower bound on the event time. In RUL applications, ignoring this kind of data can lead to biased time-to-event estimates~\citep{stepanova_censoring_2002}. Although censored data have been addressed in the literature, most existing methods for RUL prediction either require timeseries data~\citep{zhang_time-dependent_2019, wang_remaining_2022}, assume proportional hazards (PH)~\citep{wang_remaining_2022, CHEN2020101054}, or concentrate solely on estimating the probability of remaining failure-free at a future time, rather than predicting the precise time to failure~\citep{zhang_time-dependent_2019}. Additionally, evaluations of these methods are often limited, with some studies reporting only aggregated prediction results for groups of bearings~\citep{WANG2023109747, SU2021107531}, or a single prediction error across all bearings~\citep{xu_novel_2023}. To the best of our knowledge, no prior work has directly compared censoring-aware methods with non-censoring-aware methods for RUL prediction. Finally, a recurring limitation in existing works is predicting a RUL which is not monotonically decreasing over time~\citep{xu_novel_2023, LIN2024102524}, which leads to an unintuitive interpretation of the RUL, or assuming that the true RUL is a linear function of the operating hours~\citep{wang_hybrid_2020, 10409180}. Table \ref{tab:methods_comparison} shows a comparison between related work and our proposed method.

\begin{table*}[!ht]
\centering
\resizebox{\textwidth}{!}{%
\begin{tabular}{ccccc}
\toprule
Method & Approach & Advantages & Disadvantages & \makecell{Supports\\censoring} \\
\midrule
\makecell{~\cite{widodo_application_2011}} & \makecell{RVM} & \makecell{- Supports censoring\\- Gives individual event probability estimates} & \makecell{- Limited evaluation\\- Uses only KM estimates} & Yes \\
\cmidrule(lr){1-5}
\makecell{~\cite{guo_recurrent_2017}} & RNN & \makecell{- Can predict the RUL over time\\- Captures temporal dependencies} & \makecell{- Requires timeseries data\\- Does not support censoring} & No \\
\cmidrule(lr){1-5}
\makecell{~\cite{zhang_time-dependent_2019}} & \makecell{RNN/LSTM} & \makecell{- Supports censoring\\- Captures temporal dependencies} & \makecell{- Only predicts failure probability\\- Requires timeseries data} & Yes \\
\cmidrule(lr){1-5}
\makecell{~\cite{CHEN2020101054}} & \makecell{AE + CoxPH + LSTM} & \makecell{- Supports censoring\\- Captures temporal dependencies} & \makecell{- CoxPH only used to label the data\\- Assumes PH\\ - Requires timeseries data} & Yes \\
\cmidrule(lr){1-5}
\makecell{~\cite{wang_hybrid_2020}} & Fusing RVM regressors & \makecell{- Strong predictive performance\\- Gives individual RUL predictions} & \makecell{- Assumes linear RUL degradation\\- Does not support censoring} & No \\
\cmidrule(lr){1-5}
\makecell{~\cite{wang_remaining_2022}} & CoxPH + LSTM & \makecell{- Supports censoring\\- Captures temporal dependencies\\- Gives individual RUL predictions} & \makecell{- Limited evaluation\\- Assumes PH\\- No quantitative performance measurements} & Yes \\
\cmidrule(lr){1-5}
\makecell{~\cite{xu_novel_2023}} & AE + CNN-BILSTM & \makecell{- Strong predictive performance\\- Captures temporal dependencies\\- Gives individual RUL predictions} & \makecell{- Requires timeseries data\\- RUL is not monotonically decreasing\\- Assumes linear RUL degradation\\- Does not support censoring} & No \\
\cmidrule(lr){1-5}
\makecell{~\cite{10409180}} & IF-SCINET & \makecell{- Predicts the initial degradation point\\- Captures temporal dependencies\\- Gives individual RUL predictions} & \makecell{- Requires timeseries data\\- RUL is not monotonically decreasing\\- Assumes linear RUL degradation\\- Does not support censoring} & No \\
\cmidrule(lr){1-5}
\makecell{~\cite{LIN2024102524}} & CPC + Transformer & \makecell{- Novel transformer-based architecture\\- Captures temporal dependencies\\- Gives individual RUL predictions} & \makecell{- Requires timeseries data\\- RUL is not monotonically decreasing\\- Supports censoring for pretraining, not for prediction} & No \\
\cmidrule(lr){1-5}
\makecell{RULSurv (Ours)} & \makecell{KL divergence + Survival Model\\(PH and non-PH)} & \makecell{- Supports censoring\\- Gives individual RUL predictions\\- Provides monotonically decreasing RUL predictions\\- Predicts the time to failure rather than the end of life} & \makecell{- Uses manual frequency bands to detect events\\- Does not capture temporal dependencies\\} & Yes \\
\bottomrule
\end{tabular}
}
\caption{Comparative summary of the advantages and disadvantages of related works and the proposed method.}
\label{tab:methods_comparison}
\end{table*}

\section{Fundamentals}

\subsection{Elements of rolling bearings}
\label{sec:elements_of_rolling_bearings}

Mechanical bearings are used to carry axial or radial loads during rotational or oscillating motion with minimal friction. During the operation of a bearing, the rolling elements in motion are separated by a lubricant film. This is done to avoid contact with the surface asperity and to distribute the applied axial or radial load across the load zone beneath the rolling elements. Bearing health depends on a wide range of factors, such as the operating condition, operating temperature, maintenance level, among others~\citep{5620974}. Therefore, it is challenging to predict when an individual bearing is going to fail. To establish a benchmark for bearing life expectancy, a standard industry measurement is the $L_{10}$. The $L_{10}$ indicates the number of hours a group of identical bearings, running at a constant speed, will last before 10\% of them have failed~\citep{ISO281}. Approximately 90\% of industrial bearings outlive the equipment in which they are installed, and the remaining 10\% are replaced for preventive reasons. Only around 0.5\% are replaced because of actual bearing failure~\citep{skf_2017_bearing_defect}.

\subsection{Defect evolution and failure monitoring}

\begin{figure}[!ht]
\centering
\includegraphics[width=1\linewidth]{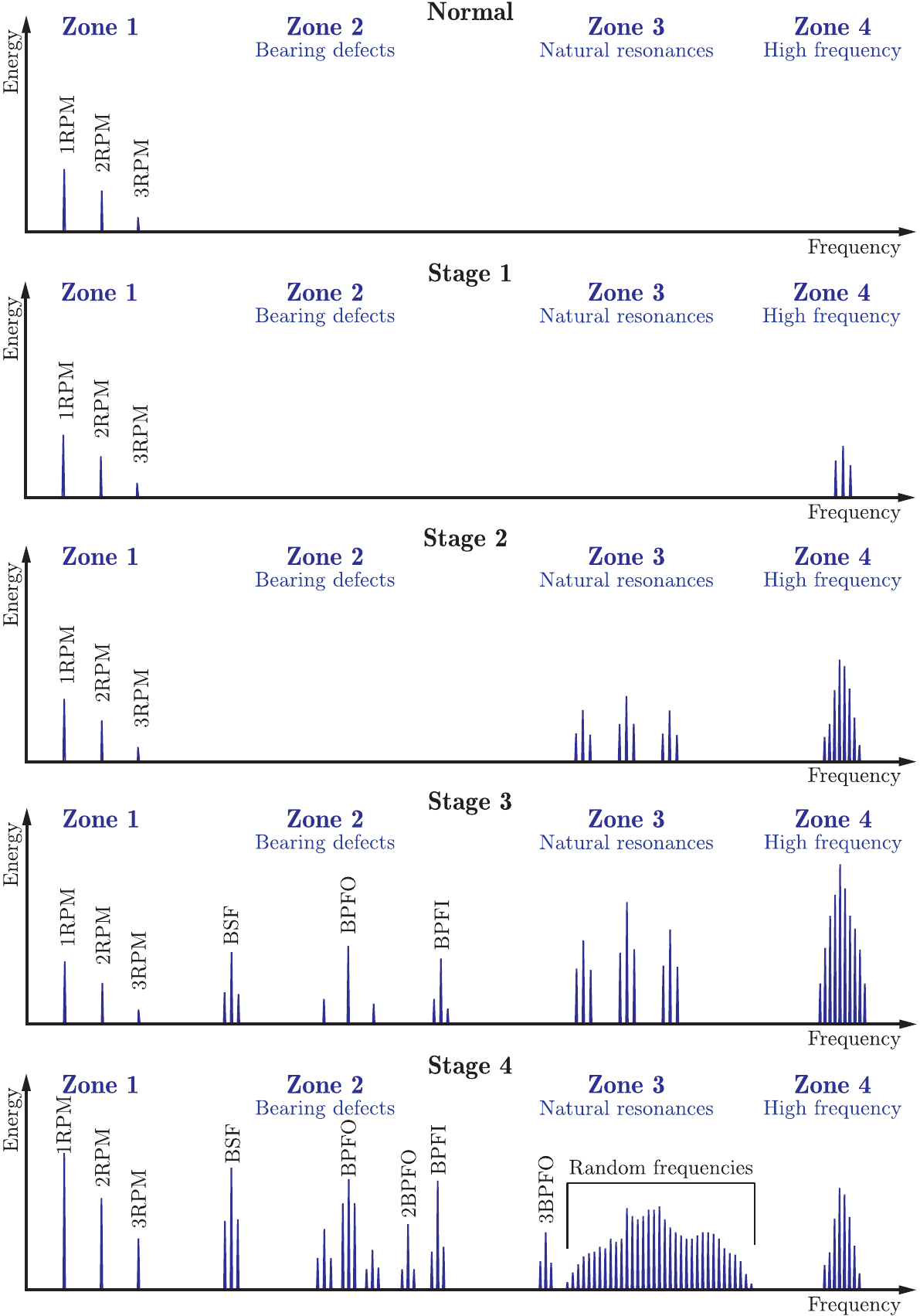}
\caption{Evolution of the frequency spectrum across the stages of a typical bearing failure. Initially, a healthy bearing exhibits frequencies solely linked to shaft phenomena, such as balance or misalignment. \textbf{Stage 1} introduces ultrasonic frequencies detectable only by specialized sensors, without visible defects on the bearing. \textbf{Stage 2} is marked by signals aligning with the bearing parts' natural resonance, alongside the initial appearance of defects upon inspection. \textbf{Stage 3} features the fundamental defect frequencies and their harmonics, modulated by the shaft speed, indicating a spread of defects. \textbf{Stage 4} is the final stage and precedes complete failure. This final stage is characterized by a mix of modulated fundamental frequencies, harmonics, and, ultimately, a shift towards a random noise floor.}
    \label{fig:defect_preogression}
\end{figure}

At the beginning of a bearing's life, operational anomalies appear minimally on the surface of the rolling elements. However, they are often characterized by sharp entry and exit edges due to material spalling and surface alteration. As a result, a large number of harmonics will often be energized during overrolling, leading to the excitation of eigenfrequencies of the bearing components that are typically in the range from 10kHz to higher. As defects progress, additional surface spalling may occur, increasing the size of the faulty area. If the overrolling continues, it will smooth the sharp edges and result in an increased amplitude of vibrations, but at a lower frequency, determined by the rate of overrolling. Figure \ref{fig:defect_preogression} illustrates the frequency spectrum of a monitored bearing through different stages of a typical bearing defect until failure occurs.

The early stages of a bearing defect can be effectively detected from weak vibration signals through the enveloped Fourier transform~\citep{randall_rolling_2011}, as this method can detect amplitude-modulated signals. By selecting an impulsive frequency band, often containing one or more of the component's eigenfrequencies and applying the Hilbert transform, the low-frequency envelope or carrier frequency can be found~\citep[Ch. 3]{randall_vibration_2021}. The carrier frequency corresponds to one of the four characteristic bearing frequencies that are bound to the bearing geometry, the number of rolling elements, and the rotation speed. 

In this work, we adopt the following frequency bands for early fault detection: the Ball Pass Frequency Outer Race (BPFO, Eq. \ref{eq:bpfo}), the Ball Pass Frequency Inner Race (BPFI, Eq. \ref{eq:bpfi}), the Ball Spin Frequency (BSF, Eq. \ref{eq:bsf}), the Fundamental Train Frequency (FTF, Eq. \ref{eq:ftf}) and the Shaft Frequency (SF). SF is obtained from the datasheet~\citep{wang_hybrid_2020}. The frequency bands are defined as:

\begin{equation}
\label{eq:bpfo}
\text{BPFO}\;(n, f_r, d, D, \phi) = \frac{n f_r}{2}\left(1-\frac{d}{D} \cos \phi\right),
\end{equation}

\begin{equation}
\label{eq:bpfi}
\text{BPFI}\;(n, f_r, d, D, \phi) = \frac{n f_r}{2}\left(1+\frac{d}{D} \cos \phi\right),
\end{equation}

\begin{equation}
\label{eq:bsf}
\text{BSF}\;(f_r, d, D, \phi) = \frac{D f_r}{2 d}\left[1-\left(\frac{d}{D} \cos \phi\right)^2\right],
\end{equation}

\begin{equation}
\label{eq:ftf}
\text{FTF}\;(f_r, d, D, \phi) = \frac{f_r}{2}\left(1-\frac{d}{D} \cos \phi\right),
\end{equation}

\noindent where $n$ is the number of rolling elements, $f_r$ is the shaft speed, $d$ is the diameter of the roller, $D$ is the mean diameter of the bearing (the center between the inner and outer ring diameter), and $\phi$ is the contact angle in degrees with respect to the radial plane. Each frequency defines what is called a critical band, $cb$. These frequency bands are critical during the evolution of a bearing failure as illustrated in Figure \ref{fig:defect_preogression}. We denote the collection of five critical bands as $CB = (cb_i)_{i=1}^{5} = (\text{BPFO}, \text{BPFI}, \text{BSF}, \text{FTF}, \text{SF})$. 

\subsection{Elements of survival analysis}
\label{sec:elements_of_survival_analysis}

We define a survival problem by a sequence of observations represented as triplets ($\bm{x}_{i}$, $t_{i}$, $\delta_{i}$), where $\bm{x}_{i} \in \mathbb{R}^{d}$ is a vector with $d$ features, $t_{i}$ denotes the time-to-event or censoring, and $\delta_{i}$ is a binary event indicator. Moreover, let $c_i$ denote the censoring time and $e_{i}$ denote the event time for the $i$th observation, thus $t_i = e_i$ if $\delta_i = 1$ or $t_i = c_i$ if $\delta_i = 0$. We consider the survival time as discrete and limit the time horizon to a finite duration, denoted $(0, ..., T_{max})$, where $T_{max}$ represents a predefined maximum time point (\eg 1 year). A survival model then estimates the probability that the event of interest occurs at time $T$ later than $t$, that is, the survival probability $S\br{t} = \P\br{T>t} = 1-\P\br{t\leq T}$. To estimate the survival probability, we can use the hazard function:

\begin{equation}\label{eq:ht}
h\br{t} = \lim_{\Delta t \rightarrow 0} \P\br{t<T\leq t+ \Delta t \vert T>t}/\Delta t\text{,}
\end{equation}

\noindent which represents the failure rate at an instant after time $t$, assuming survival beyond that time~\citep[Ch. 11]{gareth_introduction_2021}. The hazard function is related to the survival function through $h\br{t} = f\br{t}/S\br{t}$, where $f\br{t}$ is the probability density associated with $T$. Formally, $f\br{t} := \lim_{\Delta t \rightarrow 0} \P\br{t<T\leq t+\Delta t}/\Delta t $, that is, the instantaneous failure rate at time $t$. Under this definition, the function $h\br{t}$ is the probability density of $T$ conditioned on $T>t$, where the higher the value of $h\br{t}$, the higher the probability of failure. The functions $S\br{t}$, $h\br{t}$, and $f\br{t}$, are different but related ways to describe the probability distribution of $T$.

\subsection{Elements of uncertainty estimation}

Many machine learning methods for RUL prediction are indeed nonprobabilistic; they can only provide a point estimate of the RUL and disregard the inherent uncertainty associated with the prediction. The Bayesian framework offers a theoretical approach to model uncertainty when estimating the model parameters (epistemic) and when making the actual prediction (aleatoric)~\citep{magris_bayesian_2023}. This enables us to assess confidence in the predictions, which further encourages the adoption of the model as a decision support tool. \cite{8681720} introduced Bayesian Neural Networks (BNNs) for bearing health prognostics with uncertainty estimation to predict the RUL of turbofan engines. In a BNN, the weights are treated as a set of random variables $\bth \sim \P\br{\bth}$. In this context, let $\D$ denote the data, $\P\br{\D\vert \bth}$ the likelihood and $\P\br{\bth}$ the prior distribution over the parameters of interest $\bth$. Bayesian inference computes the posterior distribution $\P\br{\bth\vert \D} = \P\brth \P\br{\D\vert \bth}/\P\br{\bth}$, and then obtains the predictive distribution of unobserved data conditioned on $\D$ by marginalizing over the parameter space $\Theta$, \ie $\P\br{\bm{x}_{\text{new}}\vert \D} = \int_\Theta \P\br{\bm{x}_{\text{new}}\vert \D, \bth}\P\br{\bth \vert \D} \text{d} \bth$. However, the posterior cannot be directly computed due to $\P\br{\D}$ being intractable, and sampling methods do not scale well in high dimensions. To this end, variational inference (VI) has been proposed to approximate the posterior with a tractable parametric distribution, $q_\bz\brth$. This $q_\bz\brth$ is chosen within a class of tractable parametric distributions, $\mathcal{Q}$, by minimizing the KL divergence from $q_\bz\brth$ to $\P\br{\bth\vert \D}$. In many modern neural networks, Monte Carlo Dropout (MCD) or another timely technique is used to approximate the VI solution~\citep{gal_dropout_2016}.~\cite{lillelund_uncertainty_2023} proposed BNNs as a tool for uncertainty estimation in deep survival networks using MCD and showed a performance advantage in adopting the Bayesian framework, especially in small datasets, over nonprobabilistic models. The intrinsic probabilistic dimension of BNNs naturally allows us to make uncertainty estimates, estimate predictive errors, and plot credible intervals~\citep{qi_using_2023}. This, in turn, can justify the level of trust and confidence that practitioners have in the prediction. Nonprobabilistic models cannot give these advantages.

\begin{figure*}[!htbp]
\centering 
\includegraphics[width=1\textwidth]{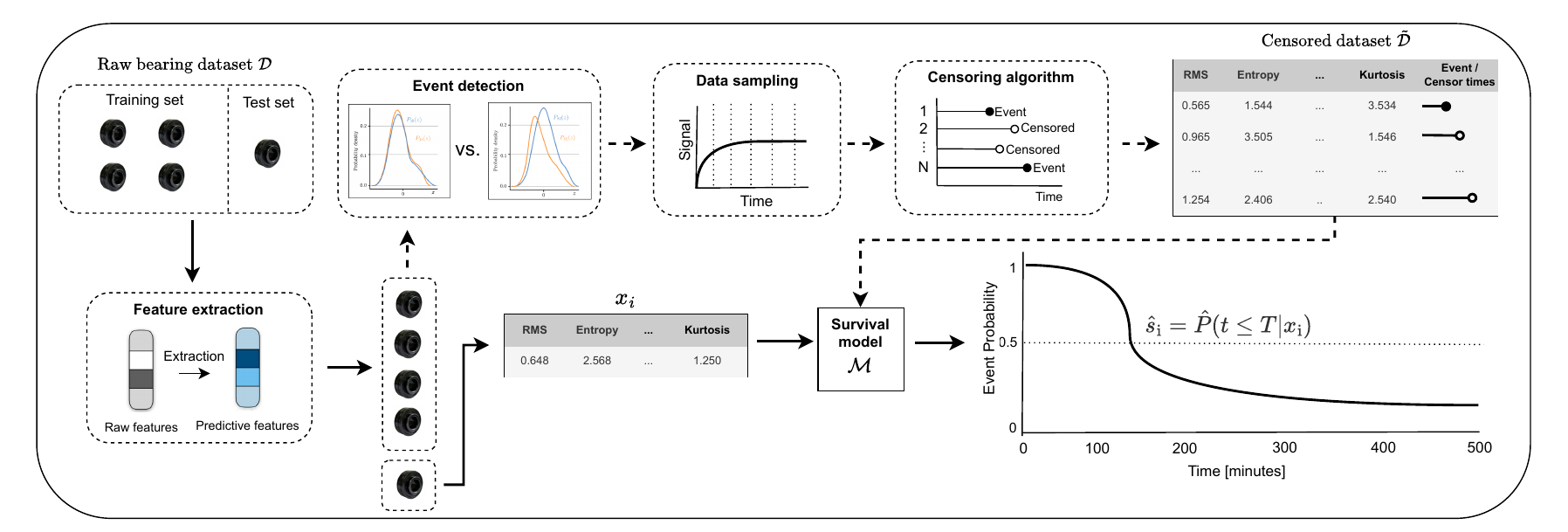}
\caption{Outline of the proposed method. Historical data for 5 ball bearings form a bearing dataset $\mathcal{D}$ (see Section \ref{sec:the_dataset}). We first perform feature extraction of all the bearings in the time-domain (see Section \ref{sec:feature_extraction}). Then, to train a survival model for RUL prediction, we designate training and test bearings and perform event detection in the frequency-domain of the training bearings (see Section \ref{sec:event_detection}). Afterwards, we convert the temporal training dataset to a supervised learning dataset by computing a rolling average (see Section \ref{sec:data_preprocessing}. We then apply random, independent censoring using the proposed censoring algorithm (see Section \ref{sec:data_censoring}), so that the event of interest is only observed for a portion of the instances. This processed dataset $\tilde{\mathcal{D}}$ thus contains $N$ instances (rows) with several time-domain features (\eg entropy, kurtosis), and the time-to-event (filled dot) or censoring (hollow dot). We use this dataset (features and event information) to train a survival model $\mathcal{M}$ (see Section \ref{sec:survival_models}) that can estimate the individual survival distribution (ISD) of the failure event, for a new test bearing $\bm{x}_{i}$, denoted as $\hat{s}_{i}$. This ISD give the probability that failure (\ie the onset of degradation) occurs after $t$ minutes post startup, for all $t>0$. It can also be used to estimate the time-to-event for this $\bm{x}_{i}$, for example, when the survival (event) curve intersects the dashed horizontal line at 50\%, which is referred to as the predicted median survival time.}
\label{fig:pipeline}
\end{figure*}

\section{Methods and materials}

This section introduces the proposed algorithm for event detection and RUL prediction in detail. Figure \ref{fig:pipeline} shows an outline of the proposed method.

\subsection{The dataset}
\label{sec:the_dataset}

We use the publicly-available Xi'an Jiaotong University (XJTU-SY)~\citep{wang_hybrid_2020} dataset to evaluate the performance of our proposed method. The dataset contains measurements of 15 deep grove ball bearings of type LDK UER204 with a dynamic load rating of 12.82 kN. Measurements are performed on a test bench, where bearings are subjected to accelerated degradation under different conditions. Three tests are performed with different loads and speeds, and five bearings are used in each test (see Tables \ref{tab:xjtu_dataset_overview} and \ref{tab:xjtu_dataset_properties}). During the test, each bearing is instrumented with two piezoelectric accelerometers of type PCB 352C33 to collect vibration signals, mounted horizontally ($X$-axis) and vertically ($Y$-axis) and positioned at a 90 degree angle to each other. Vibration signals are captured at a sampling frequency of 25.6 kHz. The sampling frequency is 25.6 kHz, and 32,768 samples (\ie 1.28 seconds of data) are recorded every 1 minute. We follow~\cite{wang_hybrid_2020} and sample only from the $X$-axis; since the load is applied horizontally, the accelerometer positioned in this direction can capture more degradation information from the tested bearings. All bearings are run to failure under very high load to accelerate degradation. This significantly increases the risk of initiating bearing defects, but also the risk of local flash heating and subsequent uncontrolled damage to a bearing.

\begin{table}[!htbp]
\begin{adjustbox}{width=\columnwidth, center}
    \begin{tabular}{@{}ccccc@{}}
    \toprule
    \multicolumn{1}{c}{\begin{tabular}[c]{@{}c@{}}Operating\\ condition\\ \end{tabular}} & \multicolumn{1}{c}{\begin{tabular}[c]{@{}c@{}}$L_{10h}$\\ \text{[h]} \end{tabular}} & \multicolumn{1}{c}{\begin{tabular}[c]{@{}c@{}}Radial force\\ \text{[kN]} \end{tabular}} & \multicolumn{1}{c}{\begin{tabular}[c]{@{}c@{}}Rotating speed\\ \text{[rpm]}\end{tabular}} & \multicolumn{1}{c}{Bearing dataset} \\ \midrule
    C1 & 9.6  & 12.0 & 2100 & Bearing 1\_1 to 1\_5 \\
    C2 & 11.7 & 11.0 & 2250 & Bearing 2\_1 to 2\_5 \\
    C3 & 14.6 & 10.0 & 2400 & Bearing 3\_1 to 3\_5 \\ \bottomrule
    \end{tabular}%
\end{adjustbox}
\caption{Overview of the XJTU-SY dataset.}
\label{tab:xjtu_dataset_overview}
\end{table}

\begin{table}[htbp]
\begin{adjustbox}{width=\columnwidth, center}
\centering
\begin{tabular}{llll}
\toprule
Property & Value & Property & Value \\
\midrule
Outer race diameter & 39.80 mm & Inner race diameter & 29.30 mm \\
Bearing mean diameter & 34.55 mm & Ball diameter & 7.92 mm \\
Number of balls & 8 & Contact angle & 0$\degree$\\
Load rating (static) & 6.65 kN & Load rating (dynamic) & 12.82 kN \\
\bottomrule
\end{tabular}
\end{adjustbox}
\caption{Properties of the LDK UER204 type bearings.}
\label{tab:xjtu_dataset_properties}
\end{table}

\subsection{Feature extraction}
\label{sec:feature_extraction}

We extract several features in the time domain from the raw bearing signal as important features for RUL prediction~\citep{tabatabaei_experimental_2020}. These features characterize the condition of mechanical systems based on sensor data. They include peak amplitude, energy, rise time, time, RMS, skewness, kurtosis, and crest factor, among others. Such features have been used before to train survival models with success~\citep{wang_remaining_2022}, and have been shown to be good predictors of the RUL~\citep{wang_hybrid_2020}. In practice, the feature data come from an accelerometer placed on the ball bearings at an angle of 90 degrees. We record a total of 12 time-domain features for our empirical analyzes, which are summarized in Table \ref{tab:dist_tab}.

\begin{table}[!ht]
\centering
\renewcommand{\arraystretch}{1.2}
\begin{tabular}{ll} 
\toprule
Feature name & Expression \\ 
\midrule
Absolute mean ($\bar{x}$) & $\frac{1}{N} \sum_{i=1}^N\left|x_i\right|$ \\
Standard deviation ($\sigma$) & $(\frac{1}{N} \sum_{i=1}^N (x_i-\mu)^2)^{1/2}$ \\
Skewness & $\frac{1}{N} \sum_{i=1}^N (x_i-\mu)^3 / \sigma^3$ \\
Kurtosis & $\frac{1}{N} \sum_{i=1}^N (x_i-\mu)^4 / \sigma^4$ \\
Entropy & $-\sum_{i=1}^N P(x_i) \log P(x_i)$ \\
Root mean square (RMS) & $(\frac{1}{N} \sum_{i=1}^N x_i^2)^{1/2}$ \\
Max value & $\max(|\bm{x}|)$  \\
Peak-To-Peak (P2P) & $\max(|\bm{x}|) - \min(|\bm{x}|)$ \\
Crest factor & $\max(|\bm{x}|) / (\frac{1}{N} \sum_{i=1}^N x_i^2)^{1/2}$ \\
Clearance factor & $\max(|\bm{x}|) / (\frac{1}{N} \sum_{i=1}^N |x_i|^{1/2})^2$ \\
Shape factor & $(\frac{1}{N} \sum_{i=1}^N x_i^2)^{1/2} / \bar{x}$ \\
Impulse & $\max(|\bm{x}|) / \bar{x}$ \\
\bottomrule
\end{tabular}
\caption{Selected time-domain features expressed in relation to the sampled signal $\bm{x} = (x)_{i=1}^N$. For convenience, $\mu$ denotes the mean value of $\bm{x}$.}
\label{tab:dist_tab}
\end{table}

\subsection{Event detection algorithm}
\label{sec:event_detection}

\begin{algorithm}[!ht]
    \caption{Proposed event detection algorithm.}
    \label{alg:EDalgo}
    \begin{algorithmic}[1]
        \Require Signal $\bm{x}$, end of life $L$, and sensitivity $\lambda_{KL}$.
        \par\noindent\rule{240pt}{0.5pt}
        \State Initialize $T=600$, $\eta=5$, $W=L-T$
        \State Initialize event times $\bm{t}= [t_1, t_2, \ldots, t_5]= L\cdot\mathbbm{1}$
        \State Compute the HT and FFT $\bm{z}=|FFT(H(\bm{x}))|$
        \ForAll{critical band, $cb$, in $CB$}
        \State Apply band-pass filter $\bm{z}_{cb} = BPF_{cb}(\bm{z})$
        \State $detected = \text{False}$
        \State $w = 0$
        \State Initialize list $\bm{\Delta_{KL}} = [\;]$
        \State Initialize $\Delta_{KL} = 0$
        \While{not $detected$ and $w \leq W$}
        \State Extract window $\bm{z}_{cb}^{w}\leftarrow\bm{z}_{cb}$
        \State Estimate PDF $P_w(z) = \text{PDF}(\bm{z}_{cb}^{w})$
        \If {$w > 0$}
        \State Calculate $KL_w$ = $\sum_{z \in Z} P_w(z) \log \frac{P_w(z)}{P_0(z)}$
        \If {$w > 1$}
        \State $\Delta_{KL}$ = $KL_{w}$ - $KL_{w-1}$
        \EndIf
        \State $\bm{\Delta_{KL}} \leftarrow \Delta_{KL}$
        \If {$w > 5$}
        \State Calculate $\sigma_{KL}$ as the SD($\bm{\Delta_{KL}}$)
        \State Calculate $THS = th_{KL}(w, \eta, \sigma_{KL}, L, \lambda_{KL})$ 
        \If {$|\Delta_{KL}| > THS$}
        \State $detected = \text{True}$
        \State $\bm{t}_{cb} = w$
        \EndIf
        \EndIf
        \EndIf
        \State $w = w + 1$
        \EndWhile
        \EndFor
        \State $t_{event} \leftarrow min(\bm{t})$
        \State \Return $t_{event}$
    \end{algorithmic}
\end{algorithm}

In this paper, we propose an event detection algorithm to estimate the time of failure before the end of life. We use this algorithm as an oracle to label the XJTU-SY dataset for model training and RUL prediction. The main motivation for our method comes from the idea that the recorded end of life, $L$, of the bearing constitutes an overestimation of the time when the bearing actually breaks. In other words, the bearing usually starts to fail or malfunction earlier than the time when the bearing becomes useless and the recording is stopped, and therefore, providing an inaccurate estimate of the breaking time is problematic because it can further bias the analysis.

In Algorithm \eqref{alg:EDalgo}, we present the pseudocode of our proposed event detection algorithm. In summary, the algorithm first performs the Hilbert transform and then applies the Fast Fourier Transform (FFT). This preliminary step is necessary to remove unwanted frequency components and eases the identification of the frequency bands of interest, as we explained in Section ~\ref{sec:elements_of_rolling_bearings}. The algorithm then systematically processes the signal obtained using a band-pass filter over the set of critical bands $CB= (cb_i)_{i=1}^{5}$, as discussed in Section~\ref{sec:elements_of_rolling_bearings} and calculated based on the specific bearing properties of the XJTU-SY data set in Table~\ref{tab:xjtu_dataset_properties}.

\begin{figure*}[!htbp]
    \centering
    \includegraphics[width=1\linewidth,trim={0, 0, 0, 0}]{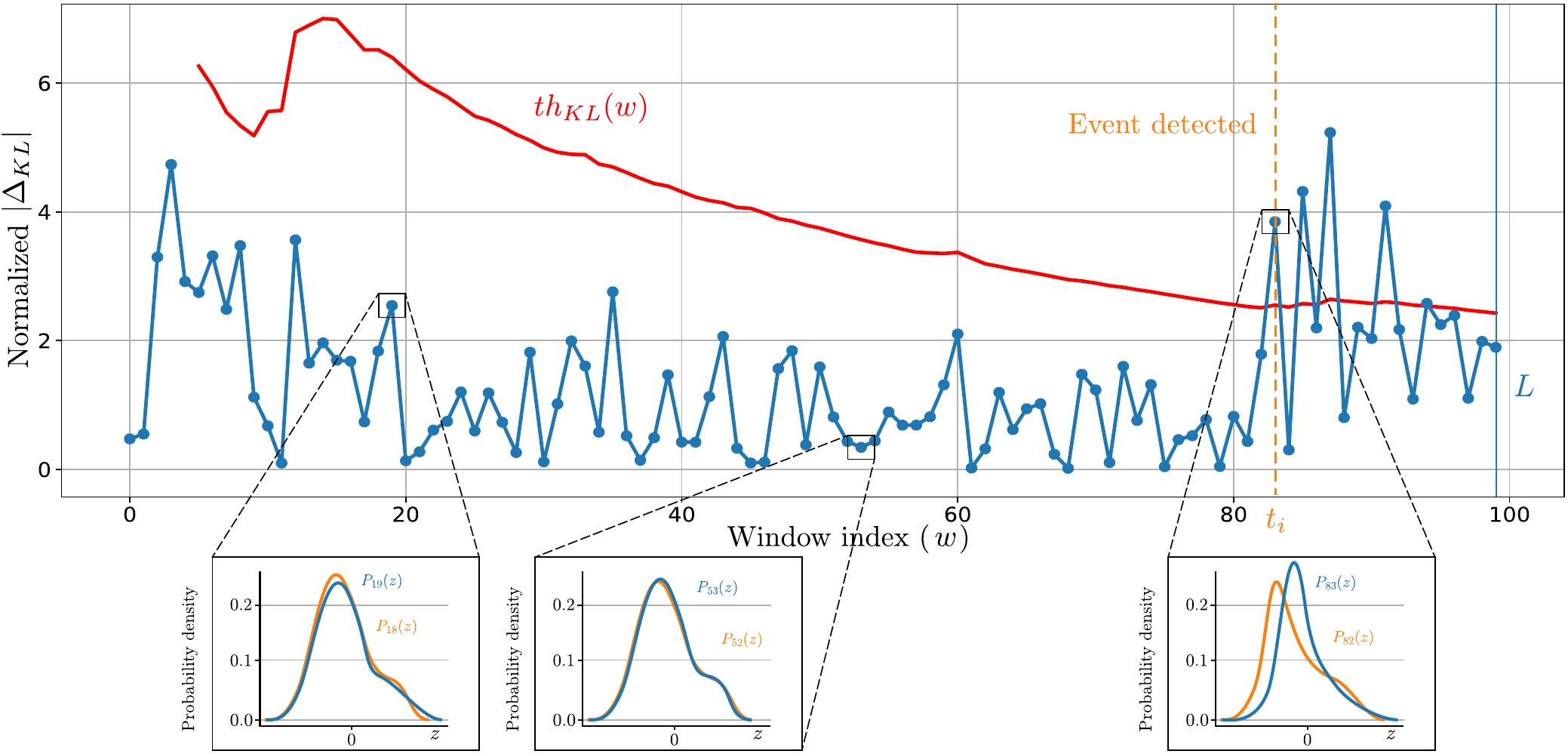}
    \caption{Visualization of the normalized $|\Delta_{KL}|$ and threshold function for a single critical band for the proposed event detection algorithm. This illustration depicts the segmentation of the signal into sequential windows ($w$), followed by the estimation of the KL divergence based on the estimated probability between consecutive windows. The diagram showcases some examples of the obtained probability density functions, and highlights the instance when the divergence exceeds the threshold, thereby indicating a probably significant malfunctioning of the bearing's performance.}
    \label{fig:event_detection}
\end{figure*}

For every critical band, the algorithm segments the observed timeseries into $W$ window frames, each with a constant duration of $T=600$ seconds. Then, the KL divergence relative to the preceding window is calculated. Following this, the algorithm tracks the variations in the differences between these values until they exceed a specific threshold. When the difference exceeds this threshold, the algorithm detects an event for the current critical band in the current window, signaling that the bearing's functioning has been compromised. For completeness, Figure \ref{fig:event_detection} visualizes the behavior of the algorithm for a particular critical band.

To control the changes in the KL divergence, we define an empirical threshold drawing inspiration from the exponential deterioration model. This threshold function is defined as:

\begin{equation}
th_{KL}(w) = \eta \cdot \sigma_{KL} \cdot \exp(-\beta t) \quad \text{with} \quad \beta = \left( \frac{1}{L} \right) \log\left( \frac{\eta}{\lambda_{KL}} \right),
\end{equation}

where $w$ represents the window index, $\beta$ is an empirical deterioration parameter with units of $1/t$, $L$ is the end of life, and $\sigma_{KL}$ is the estimated standard variation of the measured differences of the KL divergence between windows. Finally, $\eta$ and $\lambda_{KL}$ are two parameters that are linked to the natural deterioration of the bearing and bear a physical interpretation.

In particular, the parameter $\eta$ reflects the sensitivity of the approach to changes at the beginning of the recording, which is linked to the estimated standard variation, $\sigma_{KL}$. We recommend setting this parameter to $\eta = 5$ as it is sufficient to account for any initial variations until the bearing reaches a steady state. Note that this parameter may be set to higher values, yet this value is enough to avoid initial nonstationary perturbations. On the other hand, $\lambda_{KL}\sim 1$ controls how sensitive the algorithm is when it is close to the end of the bearing life. Unlike $\eta$, this additional parameter can change depending on the particular condition of the experiment, since each condition affects the bearing performance. For this experiment, we set $\lambda_{KL}=\{1.5, 1.75, 2\}$ for the high, medium, and low operating conditions, respectively.

Note that $L$ indicates the end of bearing life. In this particular dataset, it corresponds to when the bearing becomes completely useless and the recording stops. In a real industrial scenario, we do not have access to this information, since the actual time of failure is not known for a new bearing. However, we can use similar information available, such as the $L_{10}$. As detailed in Section \ref{sec:elements_of_rolling_bearings}, this value provides valuable information on the actual bearing life expectancy in a realistic scenario and can serve to fine-tune the proposed event detection algorithm.

Finally, the proposed KL-divergence metric provides better interpretability compared to deep learning architectures, such as autoencoders, because it has fewer parameters. Since our proposed method relies primarily on statistical and signal processing methods, there are no learned parameters (weights and biases) involved, making it more interpretable and parameter-efficient compared to a neural network. The primary adjustable parameters are $\eta$ and $\lambda_{KL}$, which influence the detection sensitivity, but are not learned from the data.

\subsection{Data preprocessing}
\label{sec:data_preprocessing}

The XJTU-SY dataset is a tabular dataset, where the rows represent temporal feature values, and the columns represent feature names. There are five bearings per operating condition and three operating conditions in total (see Table \ref{tab:xjtu_dataset_overview}). As a preliminary preprocessing step, we convert the temporal dataset to a supervised learning dataset by computing a rolling average with adjustable window length ($w$) and lag ($l$) based on the operating condition. For bearings under high load (C1), we use $w=2$ and $l=-1$. For medium load (C2), we use $w=4$ and $l=-3$. For the light load (C3), we use $w=6$ and $l=-5$. This reflects the length of the event horizon and means that more samples are gathered from bearings running for longer, but also that the averaging window is larger and the lag is higher. The moving window averages both the features and the event times. This produces three supervised learning datasets, $\mathcal{D}_{C1}$ ($N=635$, $d=12$), $\mathcal{D}_{C2}$ ($N=1586$, $d=12$) and $\mathcal{D}_{C3}$ ($N=4930$, $d=12$), where $N$ is the number of samples and $d$ is the number of features.

\subsection{Data censoring}
\label{sec:data_censoring}

Censored observations, \ie where the event of failure is not observed, are common in an industrial setting. Some bearings may not have failed yet at the time of observation, or they may have been decommissioned before failure occurred. Because the XJTU-SY dataset only includes complete observations, \ie the failure time is always known, we introduce random and independent censoring by a fixed percentage, $C$, to simulate censoring in an industrial setting. The amount of censoring in an industrial setting can vary and depends on several factors, \eg some bearings may be decommissioned in advance to avoid failure or perform preventive maintenance. To evaluate our method under various realistic amounts of censoring, we selected 25\%, 50\%, and 75\%, resulting in three censored data sets with the respective level of censoring. For each of these datasets, we randomly sampled the defined percentages of samples from the dataset $\mathcal{D}$, and designated those as censored. For example, to censor an observation, $i$, we calculate the time of censoring, $c_i$, as a random number between one and the observed event time, $e_i$, and use $c_i$ instead of $e_i$ as the time-to-event for observation $i$, \ie $t_i = c_i$ and $\delta_{i}=0$. This indicates that the bearing was censored at some time between when it was recorded and when it actually failed. See~\ref{app:censoring_algorithm} for a pseudocode implementation of the algorithm that automatically performs these changes.

\begin{figure*}[!ht]
\centering
\begin{subfigure}[b]{0.3\textwidth}
\centering
\includegraphics[width=1\textwidth]{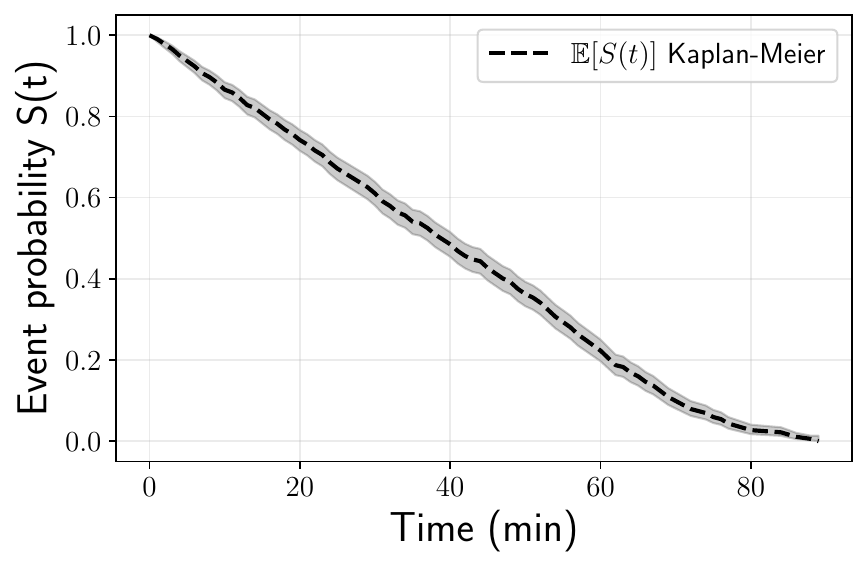}
\caption{Dataset $\mathcal{D}_{C1}$ - Censoring $25\%$}
\label{fig:kaplan_meier_C1_cens_25}
\end{subfigure}
\begin{subfigure}[b]{0.3\textwidth}
\centering
\includegraphics[width=1\textwidth]{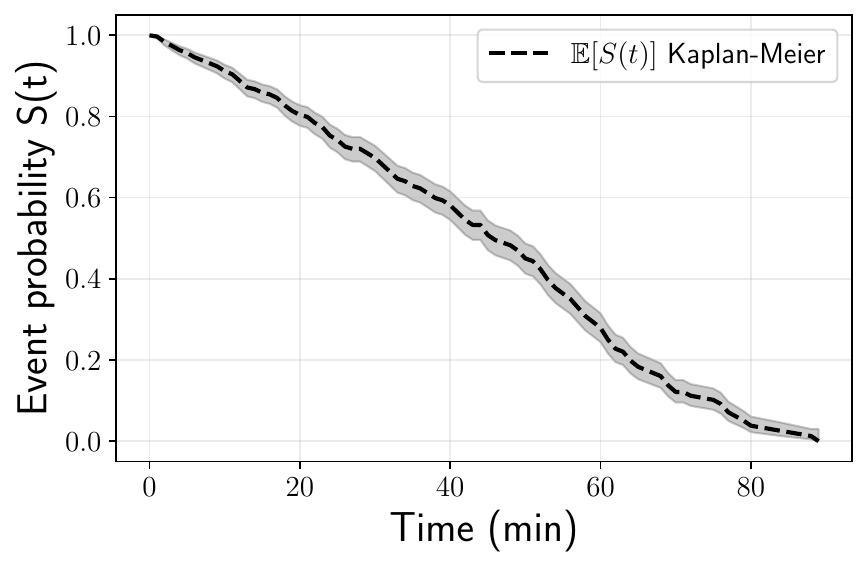}
\caption{Dataset $\mathcal{D}_{C1}$ - Censoring $50\%$}
\label{fig:kaplan_meier_C1_cens_50}
\end{subfigure}
\begin{subfigure}[b]{0.3\textwidth}
\centering
\includegraphics[width=1\textwidth]{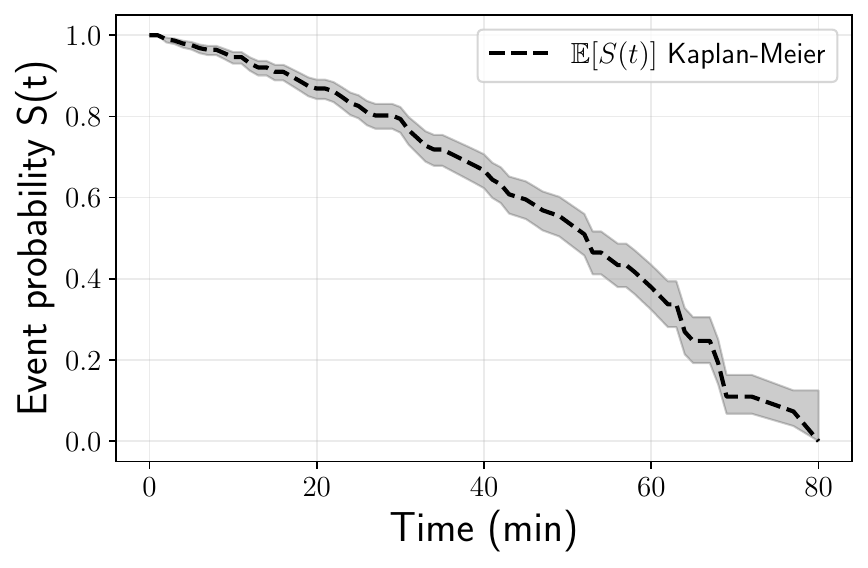}
\caption{Dataset $\mathcal{D}_{C1}$ - Censoring $75\%$}
\label{fig:kaplan_meier_C1_cens_75}
\end{subfigure}
\par\bigskip
\begin{subfigure}[b]{0.3\textwidth}
\centering
\includegraphics[width=1\textwidth]{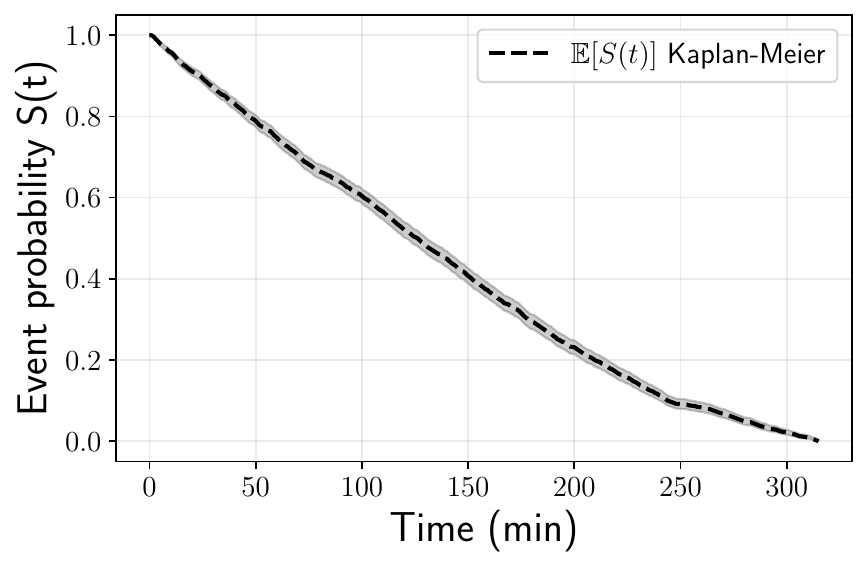}
\caption{Dataset $\mathcal{D}_{C2}$ - Censoring $25\%$}
\label{fig:kaplan_meier_C2_cens_25}
\end{subfigure}
\begin{subfigure}[b]{0.3\textwidth}
\centering
\includegraphics[width=1\textwidth]{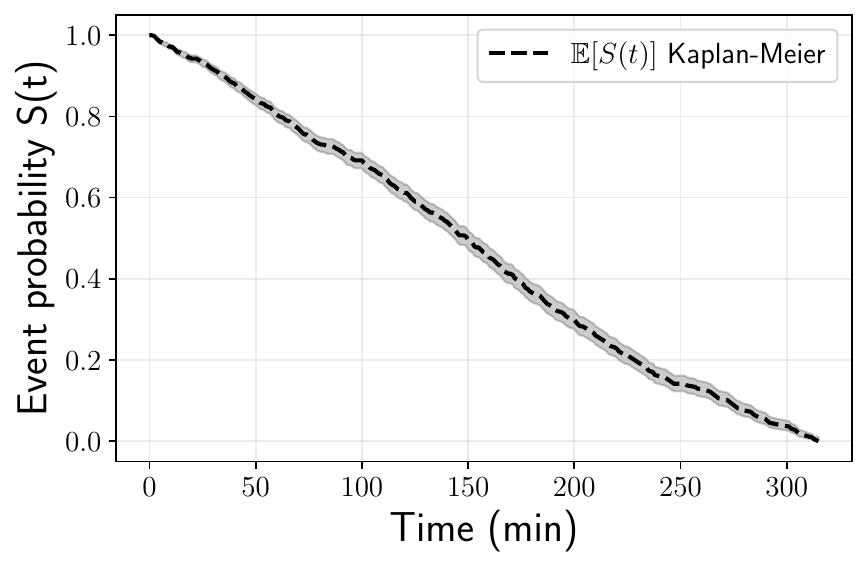}
\caption{Dataset $\mathcal{D}_{C2}$ - Censoring $50\%$}
\label{fig:kaplan_meier_C2_cens_50}
\end{subfigure}
\begin{subfigure}[b]{0.3\textwidth}
\centering
\includegraphics[width=1\textwidth]{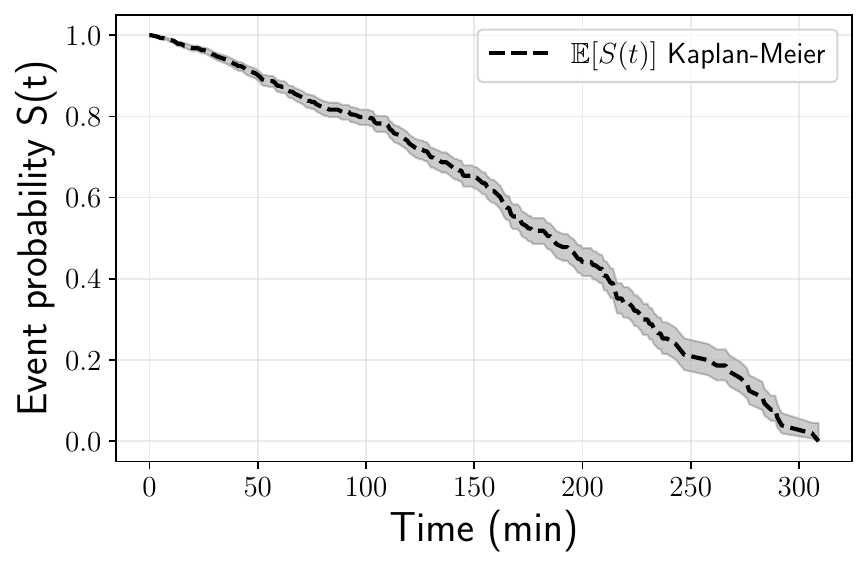}
\caption{Dataset $\mathcal{D}_{C2}$ - Censoring $75\%$}
\label{fig:kaplan_meier_C2_cens_75}
\end{subfigure}
\par\bigskip
\begin{subfigure}[b]{0.3\textwidth}
\centering
\includegraphics[width=1\textwidth]{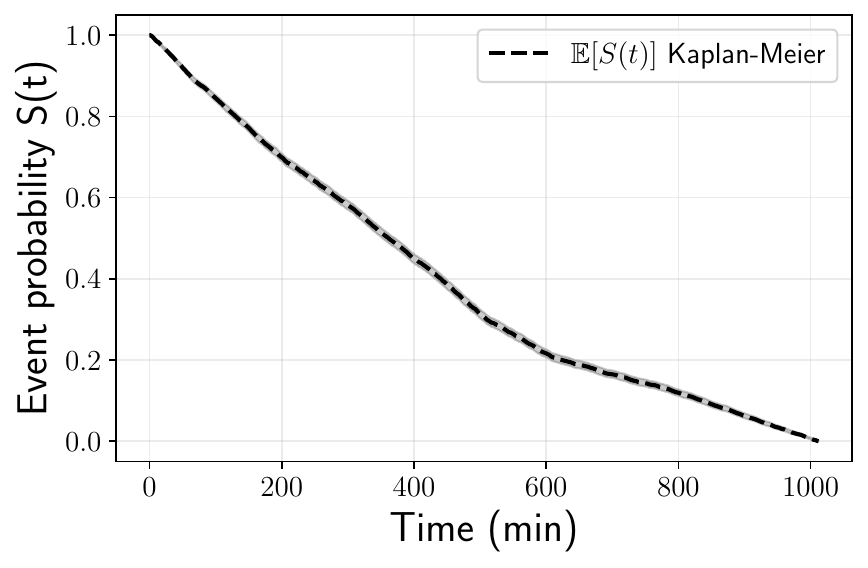}
\caption{Dataset $\mathcal{D}_{C3}$ - Censoring $25\%$}
\label{fig:kaplan_meier_C3_cens_25}
\end{subfigure}
\begin{subfigure}[b]{0.3\textwidth}
\centering
\includegraphics[width=1\textwidth]{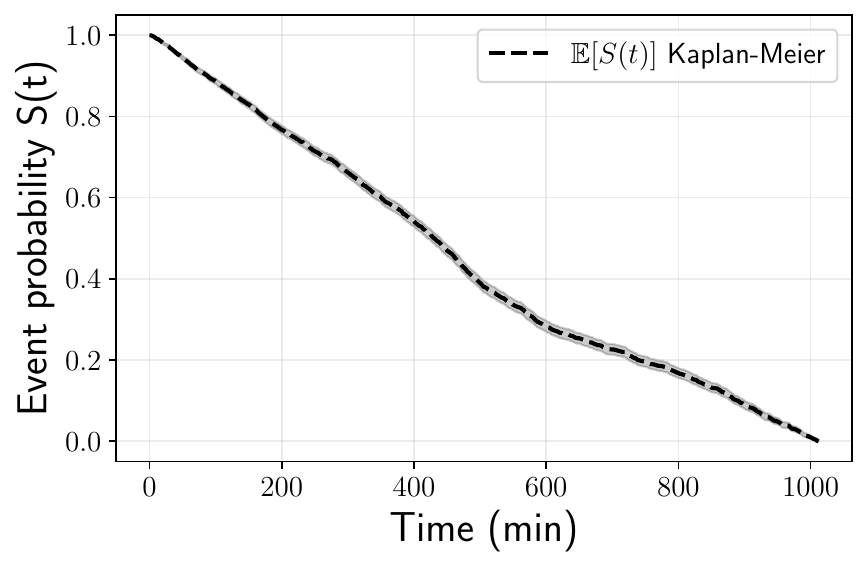}
\caption{Dataset $\mathcal{D}_{C3}$ - Censoring $50\%$}
\label{fig:kaplan_meier_C3_cens_50}
\end{subfigure}
\begin{subfigure}[b]{0.3\textwidth}
\centering
\includegraphics[width=1\textwidth]{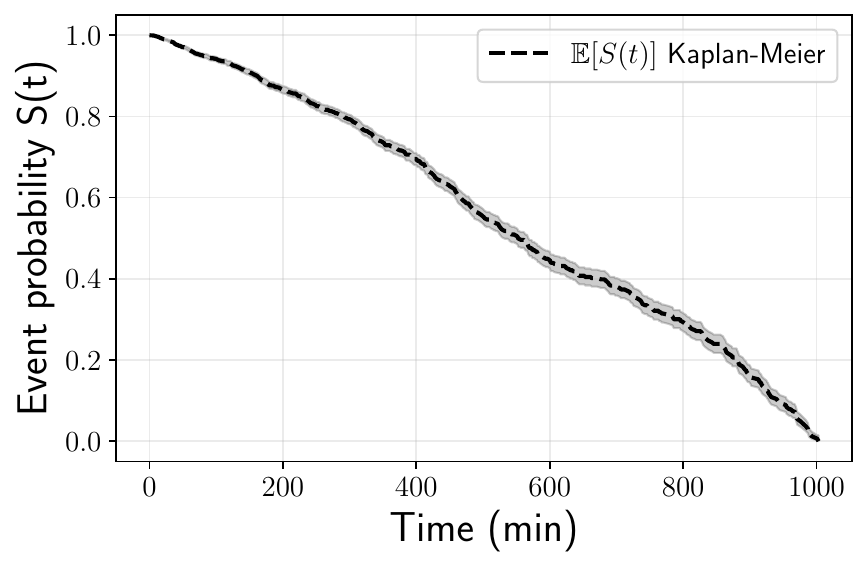}
\caption{Dataset $\mathcal{D}_{C3}$ - Censoring $75\%$}
\label{fig:kaplan_meier_C3_cens_75}
\end{subfigure}
\caption{Predicted survival probability $S(t)$ using the KM estimator under various amounts of censoring. The shaded area around the curves represent empirical 95\% confidence intervals, computed using the Greenwood formula~\citep{sawyer2003greenwood}. We see that the confidence interval increases proportionally with the level of censoring, thus indicating more uncertainty in the survival probability.}
\label{fig:kaplan_meier_censored}
\end{figure*}

Figure \ref{fig:kaplan_meier_censored} shows the KM estimate of the survival function after random censoring was applied to the individual datasets. The shaded area around the curve represents an empirical 95\% confidence interval with upper and lower bounds. These bounds are the best/worst cases of the survival function given the observed data without any parametric assumptions. We calculate the bounds using the Greenwood formula~\citep{kalbfleisch1980statistical}, and they represent the predictive uncertainty and thus the consequences of censoring when making estimates. For the upper bound on the survival function, we assume that all censored bearings do not experience the event. For the lower bound, we assume that all censored bearings immediately experience the event after they were censored. Notice that the bounds get wider as the level of censoring increases. Moreover, censoring samples at the end of the event horizon increases the predictive uncertainty more than censoring samples at the beginning of it. Generally, fewer samples are available at the end of the event horizon since most bearings have probably failed by then, so at this point, censoring is more detrimental to prediction accuracy. 

\begin{figure*}[!ht]
\centering
\begin{subfigure}[b]{0.3\textwidth}
\centering
\label{fig:event_times_C1_cens_25}
\includegraphics[width=1\textwidth]{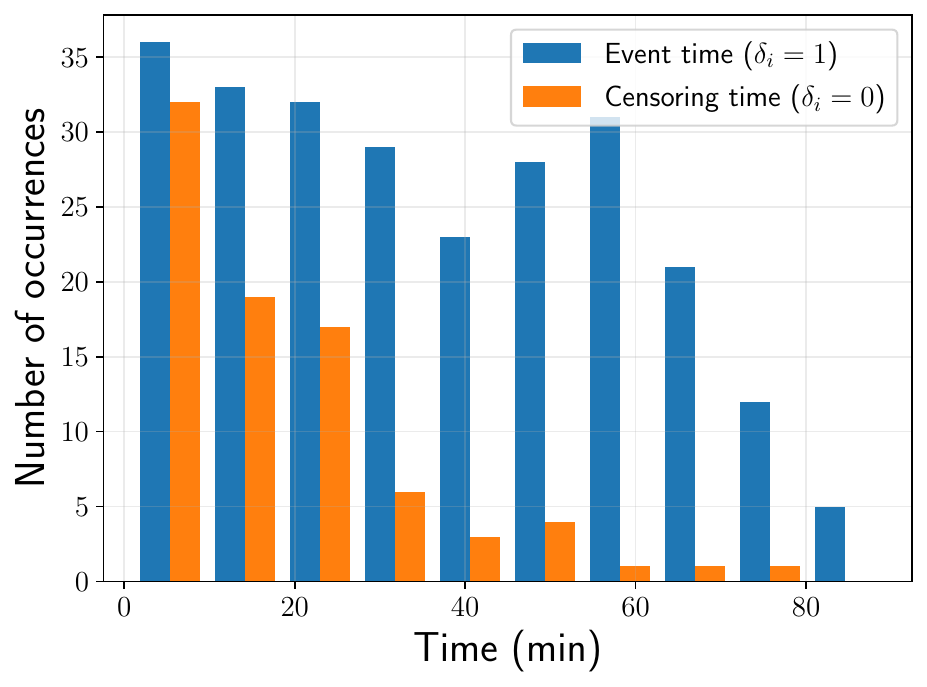}
\caption{Dataset $\mathcal{D}_{C1}$ - Censoring $25\%$}
\end{subfigure}
\begin{subfigure}[b]{0.3\textwidth}
\centering
\label{fig:event_times_C1_cens_50}
\includegraphics[width=1\textwidth]{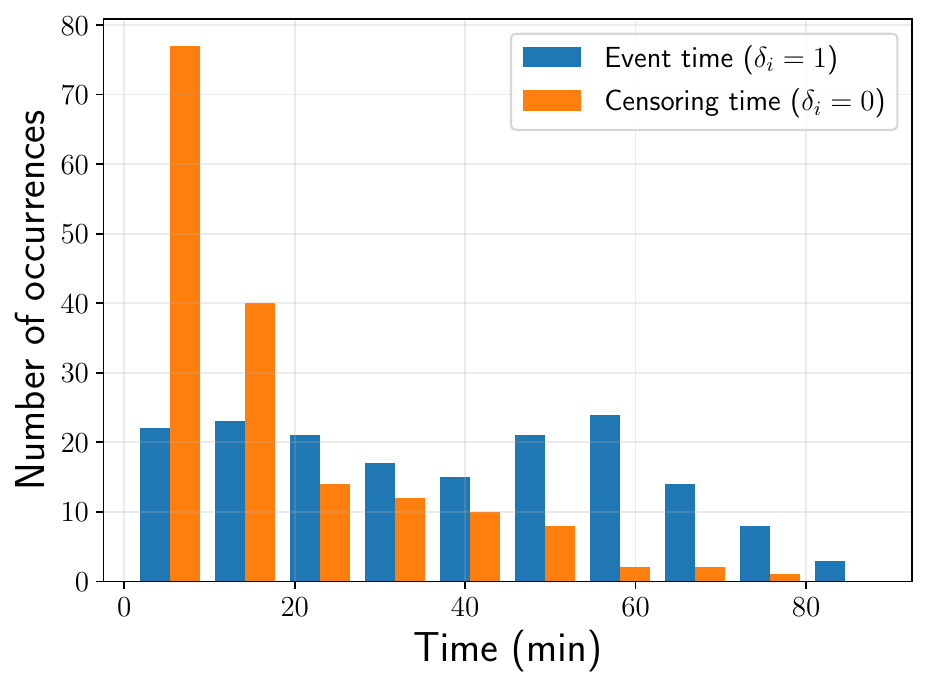}
\caption{Dataset $\mathcal{D}_{C1}$ - Censoring $50\%$}
\end{subfigure}
\begin{subfigure}[b]{0.3\textwidth}
\centering
\label{fig:event_distribution_low}
\includegraphics[width=1\textwidth]{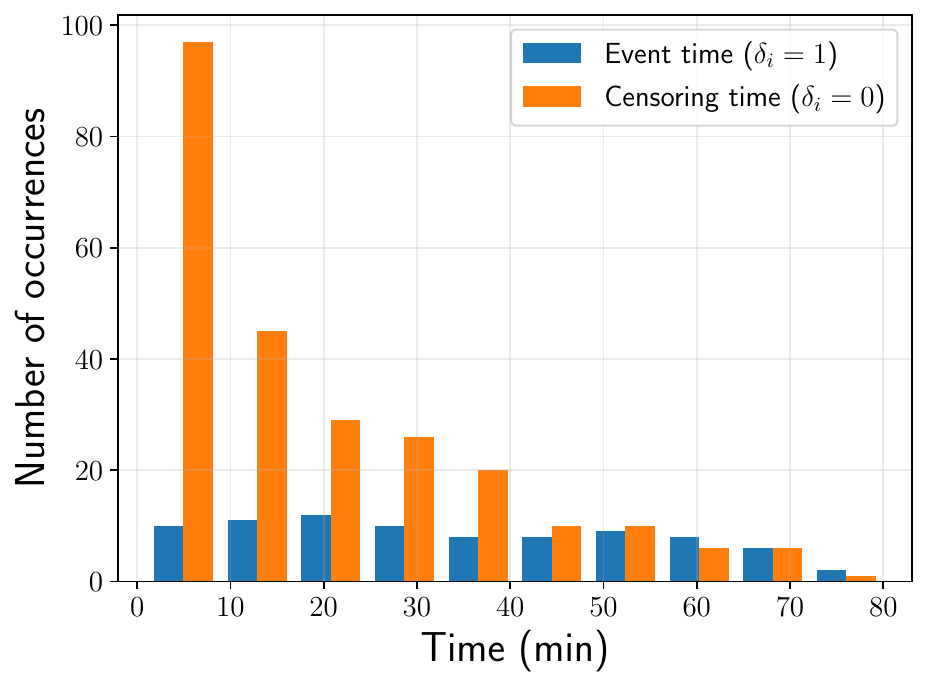}
\caption{Dataset $\mathcal{D}_{C1}$ - Censoring $75\%$}
\end{subfigure}
\par\bigskip
\begin{subfigure}[b]{0.3\textwidth}
\centering
\label{fig:event_times_C2_cens_25}
\includegraphics[width=1\textwidth]{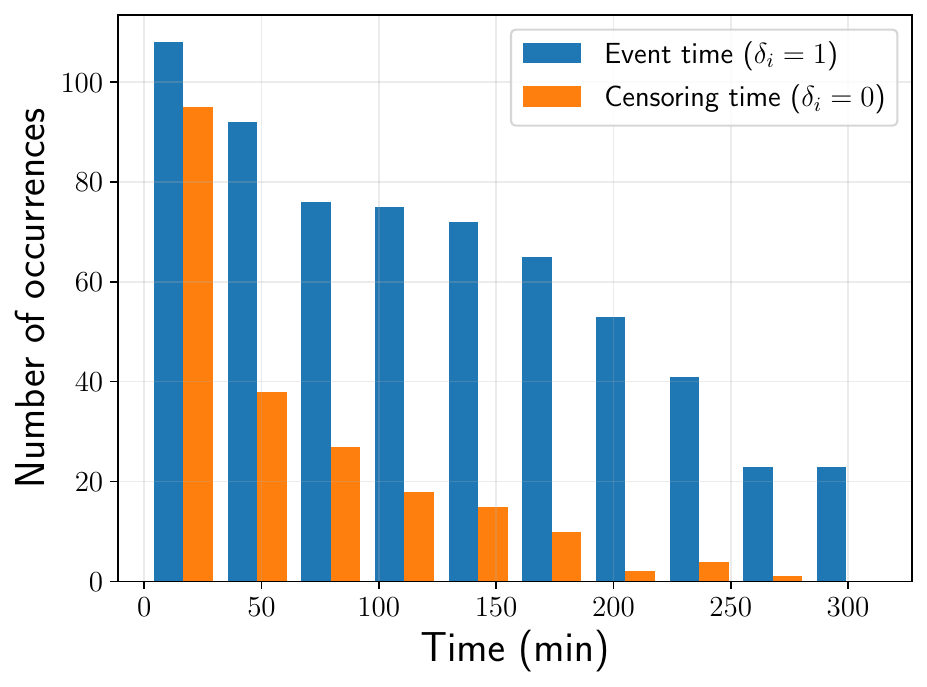}
\caption{Dataset $\mathcal{D}_{C2}$ - Censoring $25\%$}
\end{subfigure}
\begin{subfigure}[b]{0.3\textwidth}
\centering
\label{fig:event_times_C2_cens_50}
\includegraphics[width=1\textwidth]{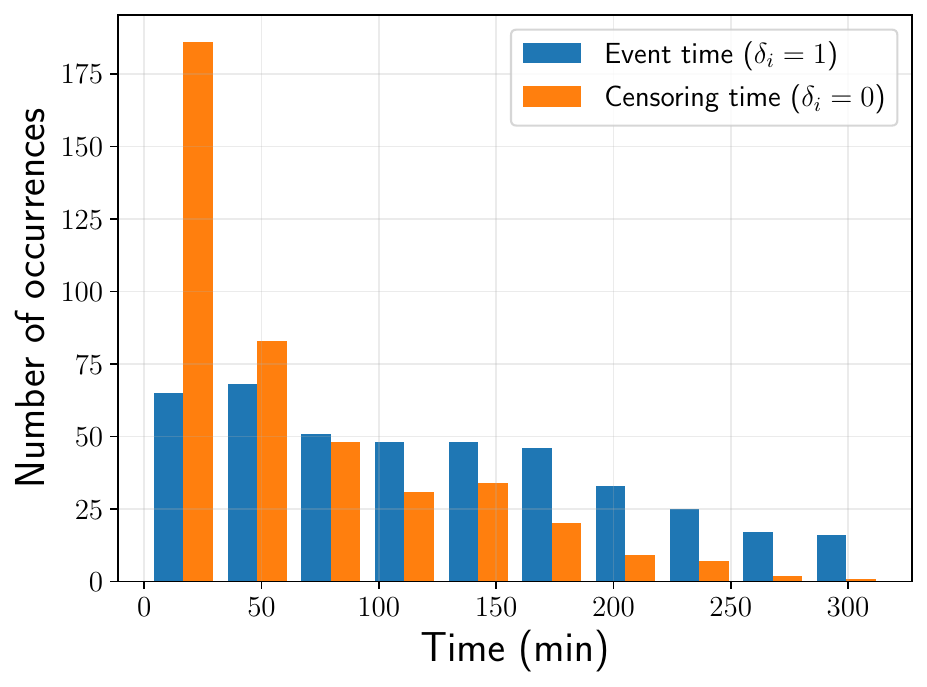}
\caption{Dataset $\mathcal{D}_{C2}$ - Censoring $50\%$}
\end{subfigure}
\begin{subfigure}[b]{0.3\textwidth}
\centering
\label{fig:event_times_C2_cens_75}
\includegraphics[width=1\textwidth]{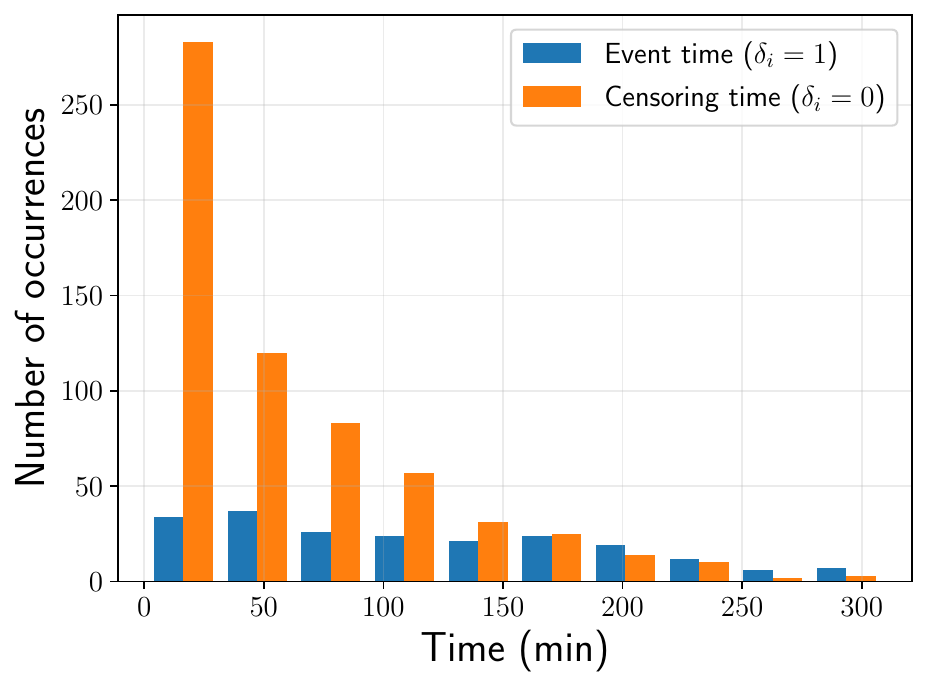}
\caption{Dataset $\mathcal{D}_{C2}$ - Censoring $75\%$}
\end{subfigure}
\par\bigskip
\begin{subfigure}[b]{0.3\textwidth}
\centering
\label{fig:event_times_C3_cens_25}
\includegraphics[width=1\textwidth]{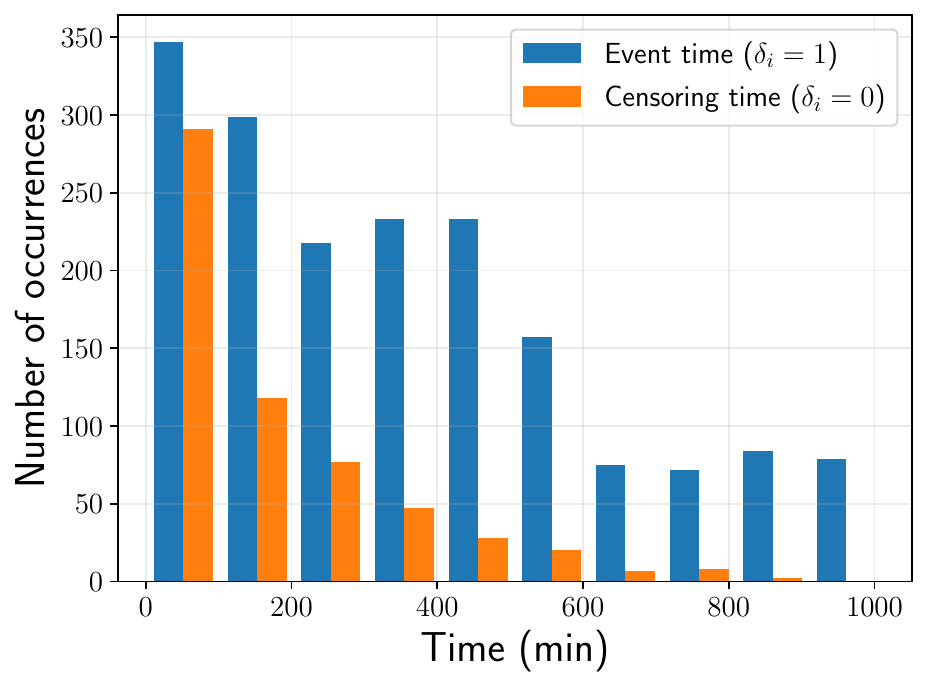}
\caption{Dataset $\mathcal{D}_{C3}$ - Censoring $25\%$}
\end{subfigure}
\begin{subfigure}[b]{0.3\textwidth}
\centering
\label{fig:event_times_C3_cens_50}
\includegraphics[width=1\textwidth]{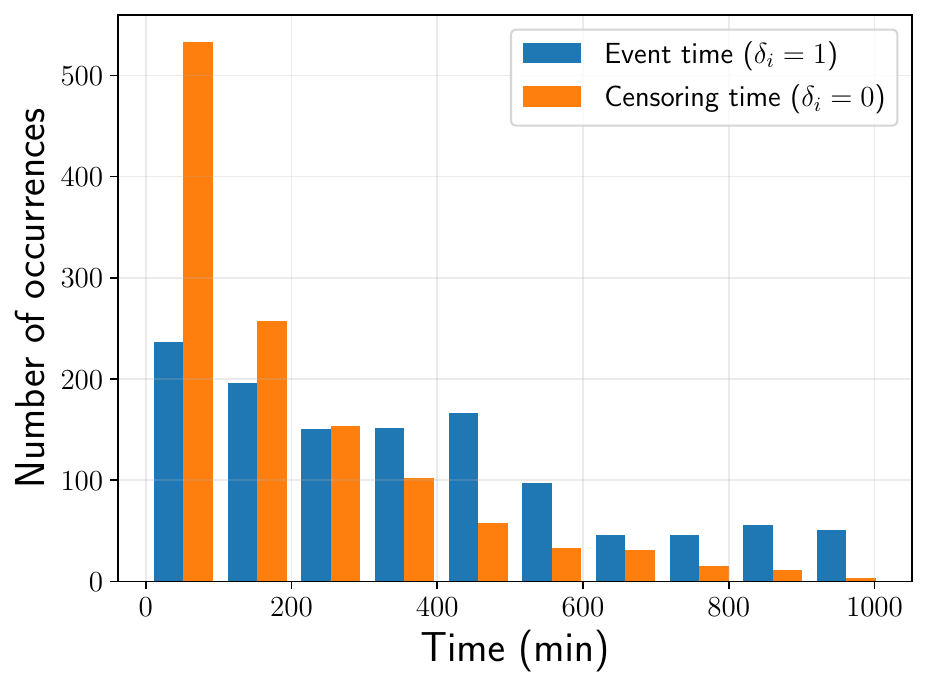}
\caption{Dataset $\mathcal{D}_{C3}$ - Censoring $50\%$}
\end{subfigure}
\begin{subfigure}[b]{0.3\textwidth}
\centering
\label{fig:event_times_C3_cens_75}
\includegraphics[width=1\textwidth]{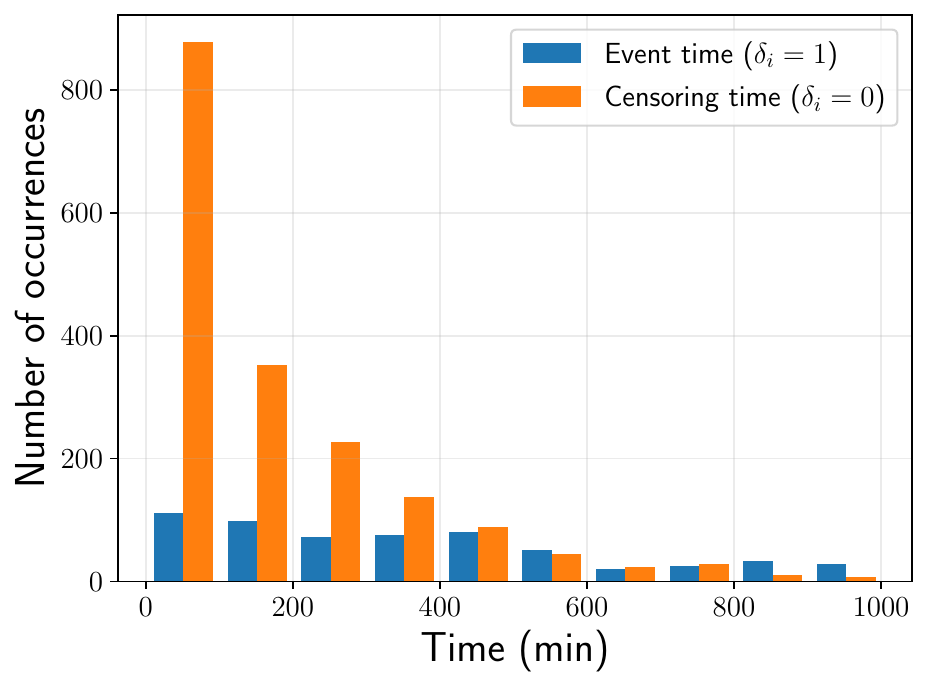}
\caption{Dataset $\mathcal{D}_{C3}$ - Censoring $75\%$}
\end{subfigure}
\caption{Event distribution of censored and uncensored event times. As censoring increases, the number of observed events decreases.}
\label{fig:event_distribution}
\end{figure*}

Figure \ref{fig:event_distribution} shows a histogram of censored and uncensored event times under various amounts of censoring.  As censoring increases, the number of observed events decreases. Under 25\% censoring, there are more observed events than censored events at each time point. Under 50\% censoring, we see more censored samples than uncensored ones at later time points, \ie between 0 and 20 minutes, since there are more samples with lower event times than higher. Under 75\% censoring, 
this phenomenon is even more pronounced.

\subsection{Survival models}
\label{sec:survival_models}

Our method consists of five survival models for RUL prediction. In this section, we will briefly describe how they work and their assumptions.

\textbf{CoxPH}~\citep{cox_regression_1972}: The Cox proportional hazards model is a widely used shallow semiparametric approach for modeling survival data. Semiparametric methods impose fewer assumptions on the survival time distribution, often yielding more robust predictions than fully-parametric methods in certain scenarios~\citep{powell_estimation_1994}. The CoxPH model defines the conditional individual hazard function as:
\begin{equation}\label{eq:hti}
h\br{t\vert \bx_i} = h_0\br{t} \cdot r_{i,\theta} = h_0\br{t} \cdot \exp \br{f\br{\bth,\bx_i}}\text{,}
\end{equation}

\noindent where $r_{i,\theta} = \exp(f\br{\bth,\bx_i})$ denotes a ``risk score'' for each instance $i$, where $f$ is a linear function and $\bth$ are the parameters. Because this model is semiparametric, estimating both the baseline hazard function and risk parameters directly from the likelihood function is challenging. Instead, training commonly relies on a partial likelihood approach that considers only the risk score:
\begin{align}\label{eq:cox_ll}
L(\bth, \D)
&= \prod_{i: \delta_i = 1} \frac{r_{\bth}(x_i)}{\sum_{j: t_j > t_i} r_{\bth}(x_j)} \
= \prod_{i: \delta_i = 1} \frac{\exp(f(\bth, x_i))}{\sum_{j: t_j > t_i} \exp(f(\bth, x_j))}\text{.}
\end{align}

The main assumption behind this model is the proportional hazards assumption, which states that the hazard ratios between instances remain constant over time. In other words, the effect of any given feature on the hazard function is multiplicative and does not vary over time.

\textbf{GBSA}~\citep{Ridgeway1999}: GBSA is a shallow survival model that adopts a gradient boosting strategy to estimate parameters $\bth$ by maximizing Eq. \ref{eq:cox_ll} and considers a flexible set of candidate features. In each boosting step, there are $q_{k}$ predetermined candidate sets of features, and for each of these $q_k$ sets, it simultaneously updates the parameter for the corresponding feature. Regular gradient boosting would update only one component of $\bth$ in component-wise boosting, or fit the gradient using all features in each step. GBSA assumes proportional hazards.

\textbf{RSF}~\citep{ishwaran_random_2008}: The Random Survival Forests model is a shallow survival model extension of decision trees for survival analysis, which are made to handle censored data. During training, data are recursively partitioned according to some splitting criterion, and similar data points based on the event of interest are placed on the same node. This ensures a decorrelation of individual trees by iteratively building each tree on a different bootstrap sample of the original training data and then at each node only evaluate the split criterion for a randomly selected subset of features and thresholds. Predictions are then made by aggregating the predictions of the individual trees in the ensemble, like the regular Random Forests model~\citep{Breiman2001}. RSF does not assume proportional hazards, but rather that sampling is representative, \ie by drawing a random number of samples, we would obtain an equal split between the number of outputs.

\textbf{MTLR}~\citep{NIPS2011_1019c809}: Multi-Task Logistic Regression is a deep survival model that directly models the individual survival function by combining multiple local logistic regression models. It splits the discretized event horizon into $K$ bins $(\tau_0 = 0,\tau_1), (\tau_1, \tau_2), \ldots, (\tau_{K-1}, \tau_K)$ and computes the probability of event at each bin by fitting a logistic regression model~\citep{qi_using_2023}. The predicted time-to-event $\hat{t_i}$ can be then be encoded deterministically using a $K$-dimensional binary sequence $\mathbf{y}_i = (y_{i,1}, y_{i,2}, \ldots, y_{i,K}) \in \mathbb{Z}^K$, where $y_{i,k} = 0$ means that the observation had not experienced the event at time $\tau_{k-1}$, and $y_{i,k} = 1$ means that the observation experienced the event before time $\tau_{k-1}$~\citep{qi_using_2023}. MTLR does not assume proportional hazards.

\textbf{BNNSurv}~\citep{lillelund_uncertainty_2023}: BNNSurv is a deep survival model that trains a Bayesian Neural Network to predict risk scores. It uses the Cox partial log-likehood to optimize the objective Eq. \eqref{eq:cox_ll}, but the weights and biases of the network are treated as random variables with an unknown posterior distribution. This allows representing the uncertainty in the parameter distribution, as opposed to deterministic models, where the weights and biases are fixed numbers. During training, they are estimated with an approximate distribution using \eg variational inference (VI), Monte Carlo dropout (MCD) or another inference technique. The output nodes thus provide the mean and standard deviation of a risk score, $r_i$, as Gaussian samples, that is, $r_i \sim \mathcal{N}\br{\mu_{\bm{x}_i},\sigma_{\bm{x}_i}}$ for a $d$-dimensional sample $\bm{x}_i$. In this work, we adopt the BNNSurv model using MCD inference and set a dropout rate of 25\%. BNNSurv is based on CoxPH and, therefore, assumes proportional hazards between the features and the risk scores.

\section{Experiments and results}

\subsection{Setup}

As described, we implement the traditional Cox model (CoxPH)~\citep{cox_regression_1972}, GBSA~\citep{Ridgeway1999}, Random Survival Forests (RSF)~\citep{ishwaran_random_2008}, and two neural network-based models, Multi-Task Logistic Regression (MTLR)~\citep{NIPS2011_1019c809} and BNNSurv using Monte-Carlo Dropout (MCD)~\citep{lillelund_uncertainty_2023}. To estimate the generalization error of each model, we arrange a stratified 5-fold cross-validation procedure, ensuring that event times and censoring levels remain consistent across all folds. For each fold, we apply $z$-score data normalization to all features and compute the event horizon on the training set. We then report the average model performance across the five folds, evaluating both predictive performance and calibration. If the respective model requires a validation set for early stopping, we allocate 30\% of the training set for this purpose. For continuous-time models, \ie CoxPH, GBSA, RSF, and BNNSurv, we compute the event horizon using the unique events and censoring times in the training set, thus the number of bins is the number of unique time points. For the discrete-time model, \ie MTLR, we calculate the square root of the number of uncensored observations, $\sqrt{\sum_{i:\delta_{i=1}}\mathbbm{1}}$, in the training set and use quantiles to evenly divide these uncensored observations into bins, as in~\cite{haider_effective_2020}.

We adopt sensible hyperparameters for the survival models based on the specific operating condition, as each condition has a distinct dataset with a different number of samples. For the baseline CoxPH model, we set the convergence criterion to $10^{-9}$ and a maximum of 100 iterations for all conditions. For the GBSA and RSF models, aligned with the three operating conditions, we set the number of estimators to $\{100, 200, 400\}$, the minimum number of samples needed to split to $\{5, 10, 20\}$, the minimum number of samples in a leaf node to $\{3, 5, 10\}$ and a maximum depth of $\{3, 5, 10\}$. For the MTLR model, we use batch sizes of $\{32, 64, 128\}$, hidden sizes of $\{16, 32, 64\}$, a fixed dropout rate of 0.25, a learning rate of $8 \times 10^{-5}$ using the Adam optimizer~\citep{KingBa15}, a penalty term of 0.01, and 5000 training epochs as suggested by~\citep{qi_using_2023}. For the BNNSurv model, we use batch and network sizes identical to MTLR, but with a learning rate of 0.001 using Adam and 100 epochs of training.

\subsection{Evaluation metrics}

To assess predictive performance, we report the mean absolute error (MAE) using hinge loss (MAE\textsubscript{H})~\citep{qi2023effective} for the censored observations, the MAE using the margin time (MAE\textsubscript{M})~\citep{qi2023effective} and the MAE using surrogate event times (MAE\textsubscript{PO}). To evaluate the calibration performance, we adopt the distribution calibration (D-Calibration)~\citep{haider_effective_2020} test and the coverage calibration (C-Calibration)~\citep{qi_using_2023} test. See \ref{app:evaluation_metrics} for more details about the evaluation metrics.

\subsection{Event detector performance}

Table \ref{tab:xjtu_ed_actual} shows a comparison between the annotated event times by the proposed event detection algorithm and the actual failure times in minutes. The average differences between the end of the recording and the estimated time of failure are 42.6\%, 48.7\%, and 56.9\% for the high, medium and low operating conditions, respectively. All events are detected before the end of the recording. These results support the use of the proposed event detector, as it provides a good estimate of the time when each bearing starts to fail (\ie the onset of degradation), but before the end of life, when the bearing physically stops working. Ideally, a failure event should be detected well in advance of the end of life for predictive maintenance to have an effect.

It is also worth noting that the bearings in the XJTU-SY dataset fail at entirely different times, even for the same operating conditions. This points to other factors -- \eg temperature, lubrication, and previous wear having a significant influence on the lifetime. We do see an increasing average error for lighter loads, which can be further mitigated by adjusting $\lambda_{KL}$ for the event detector (see Algorithm \ref{alg:EDalgo}), to ensure a proper threshold for the KL divergence. In an industrial setting, bearings are normally run under less stress than in this dataset, so the event detector would have to be recalibrated in any new dataset, keeping in mind the operational overhead predictive maintenance has, \ie what is the cost of not detecting a fault in a malfunctioning bearing (false negative), as opposed to detecting a fault in a healthy bearing (false positive).

\begin{table}[!ht]
\centering
\resizebox{\columnwidth}{!}{%
\begin{tabular}{cccccc}
\toprule
{\centering\shortstack{{Operating}\\{condition}}} &
{\centering\shortstack{{Bearing}\\{dataset}}} &
{\centering\shortstack{{$t_{i}$}\\{[m]}}} &
{\centering\shortstack{{$L$}\\{[m]}}} &
{\centering\shortstack{{Difference}\\{[m]}}} &
{\centering\shortstack{{Difference}\\{[\%]}}} \\
\midrule
\multirow{5}{*}{\shortstack{C1\\(high)}}
& Bearing 1\_1 & 77 & 122 & -45 & 36.9 \\
& Bearing 1\_2 & 89 & 160 & -71 & 44.4 \\
& Bearing 1\_3 & 62 & 157 & -95 & 60.5 \\
& Bearing 1\_4 & 70 & 121 & -51 & 42.1 \\
& Bearing 1\_5 & 36 & 51 & -15 & 29.4 \\
\cmidrule(lr){1-1}
\multirow{5}{*}{\shortstack{C2\\(medium)}}
& Bearing 2\_1 & 245 & 490 & -245 & 50.0 \\
& Bearing 2\_2 & 77 & 160 & -83 & 51.9 \\
& Bearing 2\_3 & 316 & 532 & -216 & 40.6 \\
& Bearing 2\_4 & 17 & 41 & -24 & 58.5 \\
& Bearing 2\_5 & 193 & 338 & -145 & 42.9 \\
\cmidrule(lr){1-1}
\multirow{5}{*}{\shortstack{C3\\(low)}}
& Bearing 3\_1 & 1014 & 2537 & -1523 & 60.0 \\
& Bearing 3\_2 & 612 & 2495 & -1883 & 75.5 \\
& Bearing 3\_3 & 206 & 370 & -164 & 44.3 \\
& Bearing 3\_4 & 514 & 1514 & -1000 & 66.1 \\
& Bearing 3\_5 & 69 & 113 & -44 & 38.9 \\
\bottomrule
\end{tabular}
}
\caption{Comparison of the annotated event time ($t_{i}$) and the end of life ($L$), \ie the end of the recording, as reported in the dataset~\citep{wang_hybrid_2020}, in minutes. The last two columns show the difference in minutes and the difference in relative percentages. In all cases, our algorithm detects that the bearings are malfunctioning before their end of life.}
\label{tab:xjtu_ed_actual}
\end{table}

\subsection{Survival model performance}

Table \ref{tab:xjtu_mae} shows the mean and standard deviation of the MAE, \ie the predictive performance, averaged over five cross-validation folds. Generally, we see lower MAE under high load, where the event horizon is shorter, the bearings deteriorate faster, and the event times are much more consistent (see Table \ref{tab:xjtu_ed_actual} and Figure \ref{fig:event_distribution}), compared to medium and low load. Under high load, increasing the level of censoring does not affect MAE\textsubscript{H}, but leads to worse MAE\textsubscript{M} and MAE\textsubscript{PO} results, particularly at high levels of censoring ($C=75\%$). Both MAE\textsubscript{M} and MAE\textsubscript{PO} compute pseudo-observations~\citep{andersen_pseudo-observations_2010} for censored observations, and more censoring increases the number of these synthetic observations. With more censoring comes more uncertainty, and consequently, the metric has to compute simulated event times for more samples, which generally leads to higher predictive errors. This trend is common for all models. The baseline CoxPH model gives good MAE scores under high load, but cannot keep up with its ensemble and neural network-based variants when the load decreases and the number of samples is larger, only exceeding the GBSA model. In general, the RSF model has the lowest MAE in operating conditions and in censoring levels, even outperforming neural networks, and its MAE\textsubscript{H} score is stable in all levels of censoring. However, the main trend is that censoring generally decreases predictive accuracy, as censoring leads to a loss of information, which can make it more difficult to accurately estimate the survival or the hazard function. This appears predominantly in the MAE\textsubscript{H} and MAE\textsubscript{PO} results.

Table \ref{tab:xjtu_emae} reports the eMAE for the respective datasets, models, and censoring levels. The eMAE is the absolute error between the MAE on the uncensored dataset, which in this context is called the true MAE, and the MAE with censoring. The eMAE can be defined for any MAE metric that supports censoring if an uncensored version of the data set is available, for example, $eMAE = \text{True MAE} - \text{MAE\textsubscript{H}}$, to calculate the error between the true MAE and the MAE with a hinge loss. Concerning the results, we see that MAE\textsubscript{H} and MAE\textsubscript{PO} tend to inaccurately estimate the true MAE in these datasets, while MAE\textsubscript{M} comes close to the true MAE, on average. Increasing censoring naturally makes it more difficult to calculate a result that aligns with the true MAE, which is most pronounced in the $\mathcal{D}_{C3}$ dataset.

Table \ref{tab:xjtu_true_mae} reports the true MAE for the respective datasets, models, and censoring levels. In this case, the true MAE is calculated by removing the censoring from the test set, \ie decensoring it, and keeping the censoring in the training set. Concerning the results, RSF and BNNSurv give the lowest true MAE results on average across all datasets. We also see that increasing censoring leads to worse predictive accuracy across the board. This indicates that higher levels of censoring make it more difficult for models to accurately predict the correct result, leading to larger errors. As a baseline comparison, we trained a linear model with an $\ell_{1}$ regularization term (LASSO) on uncensored observations only, \ie $\mathcal{D}^{\delta_{i=1}}$, thus disregarding censored ones. In particular, we see that disregarding the censored data leads to poorer predictive accuracy (\ie MAE), when comparing LASSO with the censoring-aware models, across the majority of datasets and censoring levels. This underlines the importance of incorporating censored observations into the model, as censored data still offer valuable information about the likelihood of an event occurring, even if the event has not yet taken place.

Table \ref{tab:xjtu_calibration} shows the calibration results in terms of D-calibration and C-calibration. To measure D-calibration, we slice the predicted survival function, $S(t)$, into 10\% intervals and accept the null hypothesis that the set $\{S_{i}(d_{i})\}$ is uniformly distributed between $[0,1]$ for all individual survival distributions $d_{i}$ if the $p$ value of a Pearson test $\chi^2$ is greater than 0.05. Thus, a calibrated survival model has $p$-values greater than 0.05. We can see from the table that all models pass the test in five out of five folds under high load, except BNNSurv, which passes the test in four out of five folds. As we decrease the load and the event horizon becomes wider, more models fail the D-calibration test. MTLR is the only model that gives D-calibrated survival curves in all datasets, regardless of the level of censoring. To measure C-calibration specifically for the BNNSurv model, we check if the predicted credible intervals match the observed probability intervals. The BNNSurv model is C-calibrated under 25\% censoring in all five folds, but the calibration performance decreases with higher censoring.

\begin{figure*}[!htbp]
\centering
\begin{subfigure}[b]{0.33\textwidth}
\centering
\includegraphics[width=1\textwidth]{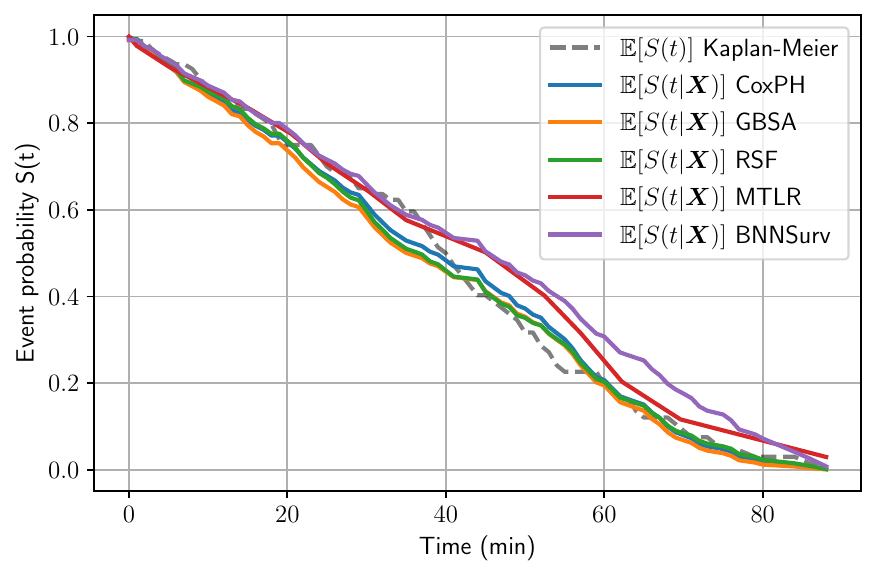}
\caption{$\mathcal{D}_{C1}$ (high).}
\label{fig:mean_survival_C1}
\end{subfigure}
\hfill
\begin{subfigure}[b]{0.33\textwidth}
\centering
\includegraphics[width=1\textwidth]{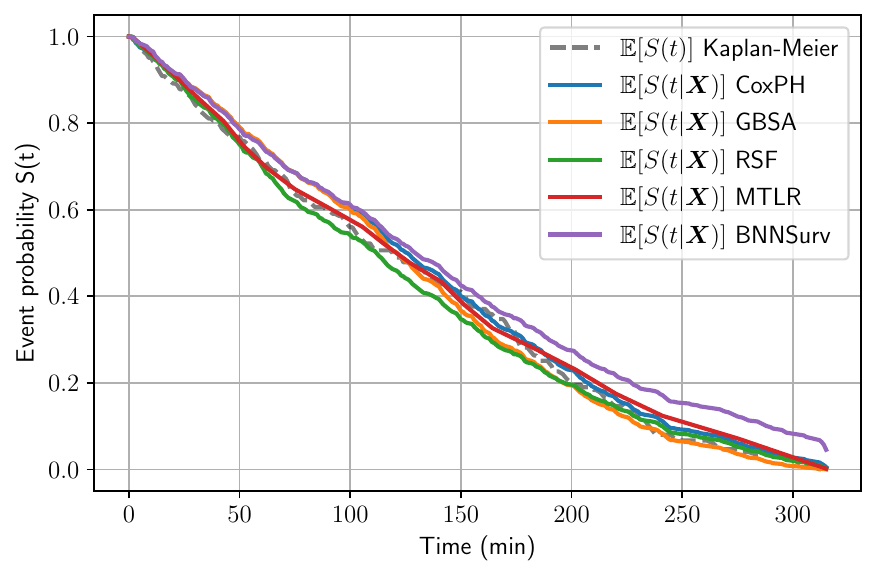}
\caption{$\mathcal{D}_{C2}$ (medium).}
\label{fig:mean_survival_C2}
\end{subfigure}
\hfill
\begin{subfigure}[b]{0.33\textwidth}
\centering
\includegraphics[width=1\textwidth]{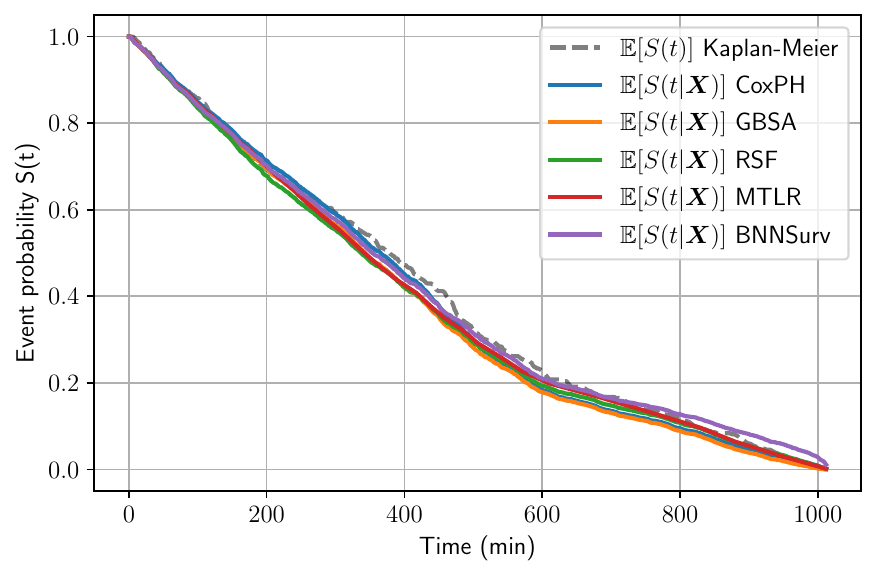}
\caption{$\mathcal{D}_{C3}$ (low).}
\label{fig:mean_survival_C3}
\end{subfigure}
\caption{Predicted mean survival curves by dataset, where 70\% of the data was used for training and 30\% was used for testing with. The censoring rate is 25\%. Solid lines: feature-based models; Dashed line: Kaplan-Meier~\citep{kaplan_nonparametric_1958} estimator). Ideally, the feature-based models should give predicted survival curves that align with the Kaplan-Meier estimator on average, which is also the case in these datasets.}
\label{fig:mean_survival_curves}
\end{figure*}

Figure \ref{fig:mean_survival_curves} shows the mean survival curves predicted by the feature-based models and the survival curve by the KM estimator. The KM estimator is a nonparametric method for estimating the survival function of a population based only on the event and censoring times. Here, we assess KM-calibration visually by observing the agreement between a model's predicted survival probabilities and the observed survival probabilities. We assume censoring to be independent, that is, bearings are censored due to reasons unrelated to the event, which means that the KM estimator is unbiased, regardless of the proportion of censoring. By visual inspection, we see that all models produce survival curves that align with the KM estimates.

\begin{table*}[!htbp]
\centering
\resizebox{\textwidth}{!}{%
\begin{tabular}{@{} cc lllllllll @{}}
\toprule
\multirow{2}{*}{\makecell{Dataset}} &
\multirow{2}{*}{\makecell{Model}} &
\multicolumn{3}{c}{25\% censoring} & \multicolumn{3}{c}{50\% censoring} & \multicolumn{3}{c}{75\% censoring} \\
\cmidrule(lr){3-5}\cmidrule(lr){6-8}\cmidrule(lr){9-11}
& & \multicolumn{1}{c}{MAE\textsubscript{H}} & \multicolumn{1}{c}{MAE\textsubscript{M}} & \multicolumn{1}{c}{MAE\textsubscript{PO}}  & \multicolumn{1}{c}{MAE\textsubscript{H}}  & \multicolumn{1}{c}{MAE\textsubscript{M}} & \multicolumn{1}{c}{MAE\textsubscript{PO}} & \multicolumn{1}{c}{MAE\textsubscript{H}} & \multicolumn{1}{c}{MAE\textsubscript{M}} & \multicolumn{1}{c}{MAE\textsubscript{PO}} \\
\midrule
\multirow{5}{*}{\makecell{$\mathcal{D}_{C1}$\\(high)\\($N=635$,\\$d=12$)}}
& CoxPH & 14.1$\pm$1.7 & 14.8$\pm$1.7 & 15.0$\pm$1.6 & 13.0$\pm$1.7 & 14.5$\pm$1.8 & 15.5$\pm$2.3 & \textbf{11.9$\pm$2.6} & \textbf{14.2$\pm$2.5} & \textbf{20.7$\pm$4.4}\\
& GBSA & 16.9$\pm$1.5 \textcolor{dimRed}{(+19.9)} & 17.7$\pm$1.5 \textcolor{dimRed}{(+19.6)} & 18.0$\pm$1.5 \textcolor{dimRed}{(+20.0)} & 16.7$\pm$1.6 \textcolor{dimRed}{(+28.5)} & 18.0$\pm$1.6 \textcolor{dimRed}{(+24.1)} & 19.1$\pm$1.9 \textcolor{dimRed}{(+23.2)} & 17.3$\pm$4.3 \textcolor{dimRed}{(+45.4)} & 18.9$\pm$4.4 \textcolor{dimRed}{(+33.1)} & 25.9$\pm$6.0 \textcolor{dimRed}{(+25.1)}\\
& RSF & \textbf{12.1$\pm$1.0} \textcolor{dimGreen}{(-14.2)} & \textbf{12.8$\pm$1.0} \textcolor{dimGreen}{(-13.5)} & \textbf{13.0$\pm$1.0} \textcolor{dimGreen}{(-13.3)} & \textbf{11.2$\pm$1.5} \textcolor{dimGreen}{(-13.8)} & \textbf{12.6$\pm$1.7} \textcolor{dimGreen}{(-13.1)} & \textbf{13.5$\pm$2.1} \textcolor{dimGreen}{(-12.9)} & 12.4$\pm$2.4 \textcolor{dimRed}{(+4.2)} & 14.6$\pm$2.1 \textcolor{dimRed}{(+2.8)} & 21.3$\pm$4.6 \textcolor{dimRed}{(+2.9)}\\
& MTLR & 13.0$\pm$1.3 \textcolor{dimGreen}{(-7.8)} & 13.7$\pm$1.4 \textcolor{dimGreen}{(-7.4)} & 13.8$\pm$1.3 \textcolor{dimGreen}{(-8.0)} & 11.9$\pm$2.6 \textcolor{dimGreen}{(-8.5)} & 13.3$\pm$2.6 \textcolor{dimGreen}{(-8.3)} & 14.2$\pm$2.9 \textcolor{dimGreen}{(-8.4)} & 13.5$\pm$2.4 \textcolor{dimRed}{(+13.4)} & 16.0$\pm$2.4 \textcolor{dimRed}{(+12.7)} & 22.6$\pm$5.0 \textcolor{dimRed}{(+9.2)}\\
& BNNSurv & 12.9$\pm$1.7 \textcolor{dimGreen}{(-8.5)} & 13.6$\pm$1.7 \textcolor{dimGreen}{(-8.1)} & 13.7$\pm$1.6 \textcolor{dimGreen}{(-8.7)} & 12.1$\pm$2.2 \textcolor{dimGreen}{(-6.9)} & 13.4$\pm$2.3 \textcolor{dimGreen}{(-7.6)} & 14.2$\pm$2.6 \textcolor{dimGreen}{(-8.4)} & 13.8$\pm$2.0 \textcolor{dimRed}{(+16.0)} & 15.8$\pm$1.8 \textcolor{dimRed}{(+11.3)} & 21.7$\pm$4.5 \textcolor{dimRed}{(+4.8)}\\
\cmidrule(lr){1-1}
\multirow{5}{*}{\makecell{$\mathcal{D}_{C2}$\\(medium)\\($N=1586$,\\$d=12$)}}
& CoxPH & 52.8$\pm$4.1 & 55.7$\pm$4.0 & 56.7$\pm$4.0 & 50.0$\pm$1.6 & 56.6$\pm$1.5 & 63.0$\pm$2.6 & 51.5$\pm$14.4 & 62.3$\pm$14.1 & 89.9$\pm$17.9\\
& GBSA & 57.4$\pm$3.1 \textcolor{dimRed}{(+8.7)} & 60.7$\pm$2.9 \textcolor{dimRed}{(+9.0)} & 61.8$\pm$2.7 \textcolor{dimRed}{(+9.0)} & 58.1$\pm$4.0 \textcolor{dimRed}{(+16.2)} & 65.2$\pm$3.7 \textcolor{dimRed}{(+15.2)} & 71.9$\pm$4.7 \textcolor{dimRed}{(+14.1)} & 61.6$\pm$6.8 \textcolor{dimRed}{(+19.6)} & 71.5$\pm$7.1 \textcolor{dimRed}{(+14.8)} & 100.0$\pm$10.7 \textcolor{dimRed}{(+11.2)}\\
& RSF & \textbf{29.4$\pm$2.2} \textcolor{dimGreen}{(-44.3)} & \textbf{32.7$\pm$2.4} \textcolor{dimGreen}{(-41.3)} & \textbf{33.4$\pm$2.6} \textcolor{dimGreen}{(-41.1)} & \textbf{29.8$\pm$4.4} \textcolor{dimGreen}{(-40.4)} & \textbf{36.8$\pm$4.0} \textcolor{dimGreen}{(-35.0)} & \textbf{42.5$\pm$4.9} \textcolor{dimGreen}{(-32.5)} & \textbf{31.8$\pm$7.3} \textcolor{dimGreen}{(-38.3)} & \textbf{43.1$\pm$7.0} \textcolor{dimGreen}{(-30.8)} & \textbf{69.9$\pm$13.3} \textcolor{dimGreen}{(-22.2)}\\
& MTLR & 47.0$\pm$5.4 \textcolor{dimGreen}{(-11.0)} & 49.9$\pm$5.2 \textcolor{dimGreen}{(-10.4)} & 50.7$\pm$5.0 \textcolor{dimGreen}{(-10.6)} & 45.5$\pm$2.6 \textcolor{dimGreen}{(-9.0)} & 52.3$\pm$3.0 \textcolor{dimGreen}{(-7.6)} & 58.2$\pm$3.8 \textcolor{dimGreen}{(-7.6)} & 42.8$\pm$6.3 \textcolor{dimGreen}{(-16.9)} & 54.5$\pm$6.5 \textcolor{dimGreen}{(-12.5)} & 81.6$\pm$12.2 \textcolor{dimGreen}{(-9.2)}\\
& BNNSurv & 46.1$\pm$4.8 \textcolor{dimGreen}{(-12.7)} & 48.9$\pm$4.6 \textcolor{dimGreen}{(-12.2)} & 49.8$\pm$4.6 \textcolor{dimGreen}{(-12.2)} & 43.5$\pm$1.2 \textcolor{dimGreen}{(-13.0)} & 50.1$\pm$1.0 \textcolor{dimGreen}{(-11.5)} & 56.4$\pm$1.7 \textcolor{dimGreen}{(-10.5)} & 40.0$\pm$5.9 \textcolor{dimGreen}{(-22.3)} & 52.9$\pm$6.4 \textcolor{dimGreen}{(-15.1)} & 81.3$\pm$12.6 \textcolor{dimGreen}{(-9.6)}\\
\cmidrule(lr){1-1}
\multirow{5}{*}{\makecell{$\mathcal{D}_{C3}$\\(low)\\($N=4930$,\\$d=12$)}}
& CoxPH & 173.9$\pm$8.4 & 185.9$\pm$7.8 & 189.8$\pm$7.9 & 164.6$\pm$10.2 & 192.2$\pm$8.7 & 217.7$\pm$7.8 & 196.8$\pm$27.1 & 236.1$\pm$24.5 & 330.3$\pm$35.5\\
& GBSA & 188.7$\pm$3.2 \textcolor{dimRed}{(+8.5)} & 202.3$\pm$2.9 \textcolor{dimRed}{(+8.8)} & 206.3$\pm$3.3 \textcolor{dimRed}{(+8.7)} & 189.2$\pm$9.6 \textcolor{dimRed}{(+14.9)} & 218.8$\pm$7.8 \textcolor{dimRed}{(+13.8)} & 245.1$\pm$6.3 \textcolor{dimRed}{(+12.6)} & 205.6$\pm$11.1 \textcolor{dimRed}{(+4.5)} & 246.5$\pm$11.0 \textcolor{dimRed}{(+4.4)} & 342.6$\pm$20.5 \textcolor{dimRed}{(+3.7)}\\
& RSF & \textbf{98.8$\pm$7.3} \textcolor{dimGreen}{(-43.2)} & \textbf{110.2$\pm$6.6} \textcolor{dimGreen}{(-40.7)} & \textbf{113.1$\pm$7.5} \textcolor{dimGreen}{(-40.4)} & \textbf{96.7$\pm$6.2} \textcolor{dimGreen}{(-41.3)} & \textbf{123.6$\pm$5.2} \textcolor{dimGreen}{(-35.7)} & \textbf{146.6$\pm$5.9} \textcolor{dimGreen}{(-32.7)} & \textbf{96.3$\pm$12.9} \textcolor{dimGreen}{(-51.1)} & \textbf{141.9$\pm$13.7} \textcolor{dimGreen}{(-39.9)} & \textbf{230.8$\pm$23.1} \textcolor{dimGreen}{(-30.1)}\\
& MTLR & 143.4$\pm$7.7 \textcolor{dimGreen}{(-17.5)} & 155.4$\pm$7.1 \textcolor{dimGreen}{(-16.4)} & 158.6$\pm$7.1 \textcolor{dimGreen}{(-16.4)} & 139.2$\pm$8.2 \textcolor{dimGreen}{(-15.4)} & 166.9$\pm$7.1 \textcolor{dimGreen}{(-13.2)} & 190.2$\pm$6.9 \textcolor{dimGreen}{(-12.6)} & 137.2$\pm$15.4 \textcolor{dimGreen}{(-30.3)} & 182.2$\pm$14.6 \textcolor{dimGreen}{(-22.8)} & 267.9$\pm$25.4 \textcolor{dimGreen}{(-18.9)}\\
& BNNSurv & 132.4$\pm$6.9 \textcolor{dimGreen}{(-23.9)} & 144.2$\pm$6.6 \textcolor{dimGreen}{(-22.4)} & 147.9$\pm$7.1 \textcolor{dimGreen}{(-22.1)} & 125.5$\pm$9.3 \textcolor{dimGreen}{(-23.8)} & 155.0$\pm$8.0 \textcolor{dimGreen}{(-19.4)} & 180.5$\pm$7.4 \textcolor{dimGreen}{(-17.1)} & 121.4$\pm$9.9 \textcolor{dimGreen}{(-38.3)} & 178.8$\pm$9.4 \textcolor{dimGreen}{(-24.3)} & 274.9$\pm$20.9 \textcolor{dimGreen}{(-16.8)}\\
\bottomrule
\end{tabular}
}
\caption{Mean $\pm$ standard deviation of three MAE metrics under censoring, averaged over five folds, with relative improvement/degradation compared to the baseline CoxPH model. Lower values of MAE indicate better predictive accuracy. In many cases, the RSF model gives notable improvements in mean absolute error over the baseline CoxPH and neural network-based models. $N$: sample size, $d$: number of features.}
\label{tab:xjtu_mae}
\end{table*}

\begin{table*}[!htbp]
\centering
\resizebox{\textwidth}{!}{%
\begin{tabular}{@{} cc lllllllll @{}}
\toprule
\multirow{2}{*}{\makecell{Dataset}} &
\multirow{2}{*}{\makecell{Model}} &
\multicolumn{3}{c}{25\% censoring} & \multicolumn{3}{c}{50\% censoring} & \multicolumn{3}{c}{75\% censoring} \\
\cmidrule(lr){3-5}\cmidrule(lr){6-8}\cmidrule(lr){9-11}
& & \multicolumn{1}{c}{eMAE\textsubscript{H}}  & \multicolumn{1}{c}{eMAE\textsubscript{M}} & \multicolumn{1}{c}{eMAE\textsubscript{PO}}  & \multicolumn{1}{c}{eMAE\textsubscript{H}} & \multicolumn{1}{c}{eMAE\textsubscript{M}}  & \multicolumn{1}{c}{eMAE\textsubscript{PO}} & \multicolumn{1}{c}{eMAE\textsubscript{H}} & \multicolumn{1}{c}{eMAE\textsubscript{M}} & \multicolumn{1}{c}{eMAE\textsubscript{PO}} \\
\midrule
\multirow{5}{*}{\makecell{$\mathcal{D}_{C1}$ (high)\\($N=635$,\\$d=12$)}}
& CoxPH & 0.8$\pm$0.4 & 0.6$\pm$0.4& 0.6$\pm$0.5 & \textbf{2.5$\pm$0.9} & \textbf{1.3$\pm$0.6} & 1.1$\pm$0.8 & 5.5$\pm$2.8 & 3.9$\pm$1.3& 5.0$\pm$2.5\\
& GBSA & 0.7$\pm$0.4 \textcolor{dimGreen}{(-12.5)} & 0.5$\pm$0.4 \textcolor{dimGreen}{(-16.7)}& 0.6$\pm$0.5 \textcolor{dimGreen}{(0.0)} & 2.7$\pm$0.8 \textcolor{dimRed}{(+8.0)} & 1.5$\pm$0.8 \textcolor{dimRed}{(+15.4)}& \textbf{0.9$\pm$0.4} \textcolor{dimGreen}{(-18.2)} & 6.8$\pm$2.0 \textcolor{dimRed}{(+23.6)} & 5.9$\pm$0.7 \textcolor{dimRed}{(+51.3)}& 5.3$\pm$2.6 \textcolor{dimRed}{(+6.0)}\\
& RSF & \textbf{0.6$\pm$0.5} \textcolor{dimGreen}{(-25.0)} & \textbf{0.5$\pm$0.2} \textcolor{dimGreen}{(-16.7)}& 0.6$\pm$0.2 \textcolor{dimGreen}{(0.0)} & 2.9$\pm$0.7 \textcolor{dimRed}{(+16.0)} & 1.4$\pm$0.8 \textcolor{dimRed}{(+7.7)}& 1.2$\pm$0.4 \textcolor{dimRed}{(+9.1)} & 4.8$\pm$2.2 \textcolor{dimGreen}{(-12.7)} & 3.0$\pm$1.0 \textcolor{dimGreen}{(-23.1)}& \textbf{4.4$\pm$3.4} \textcolor{dimGreen}{(-12.0)}\\
& MTLR & 0.8$\pm$0.6 \textcolor{dimGreen}{(0.0)} & 0.6$\pm$0.3 \textcolor{dimGreen}{(0.0)}& \textbf{0.6$\pm$0.1} \textcolor{dimGreen}{(0.0)} & 2.7$\pm$0.8 \textcolor{dimRed}{(+8.0)} & 1.4$\pm$0.7 \textcolor{dimRed}{(+7.7)}& 1.3$\pm$0.5 \textcolor{dimRed}{(+18.2)} & \textbf{4.2$\pm$2.0} \textcolor{dimGreen}{(-23.6)} & \textbf{2.6$\pm$0.6} \textcolor{dimGreen}{(-33.3)}& 5.6$\pm$3.3 \textcolor{dimRed}{(+12.0)}\\
& BNNSurv & 1.3$\pm$0.9 \textcolor{dimRed}{(+62.5)} & 0.9$\pm$0.7 \textcolor{dimRed}{(+28.6)}& 0.9$\pm$0.6 \textcolor{dimRed}{(+50.0)} & 3.2$\pm$0.9 \textcolor{dimRed}{(+33.3)} & 2.0$\pm$0.9 \textcolor{dimRed}{(+53.8)}& 1.7$\pm$0.8 \textcolor{dimRed}{(+41.7)} & 4.7$\pm$2.7 \textcolor{dimGreen}{(-14.5)} & 3.5$\pm$1.2 \textcolor{dimGreen}{(-10.3)}& \textbf{4.4$\pm$3.4} \textcolor{dimGreen}{(-12.0)}\\
\cmidrule(lr){1-1}
\multirow{5}{*}{\makecell{$\mathcal{D}_{C2}$ (medium)\\($N=1586$,\\$d=12$)}}
& CoxPH & 3.5$\pm$1.6 & 1.6$\pm$0.6& 1.4$\pm$0.7 & 7.1$\pm$1.5 & \textbf{1.2$\pm$1.0} & 5.9$\pm$0.7 & 18.9$\pm$3.7 & 8.1$\pm$3.8& 19.6$\pm$7.0\\
& GBSA & 3.3$\pm$0.6 \textcolor{dimGreen}{(-5.7)} & \textbf{0.4$\pm$0.2} \textcolor{dimGreen}{(-75.0)}& \textbf{1.0$\pm$0.7} \textcolor{dimGreen}{(-28.6)} & 8.9$\pm$3.7 \textcolor{dimRed}{(+25.4)} & 2.8$\pm$2.5 \textcolor{dimRed}{(+133.3)}& 5.6$\pm$3.0 \textcolor{dimGreen}{(-5.1)} & 20.3$\pm$1.1 \textcolor{dimRed}{(+7.4)} & 10.4$\pm$1.2 \textcolor{dimRed}{(+28.4)}& \textbf{18.1$\pm$7.9} \textcolor{dimGreen}{(-7.7)}\\
& RSF & \textbf{2.1$\pm$0.6} \textcolor{dimGreen}{(-40.0)} & 1.3$\pm$0.7 \textcolor{dimGreen}{(-18.8)}& 1.9$\pm$1.1 \textcolor{dimRed}{(+35.7)} & \textbf{4.5$\pm$1.3} \textcolor{dimGreen}{(-36.6)} & 2.5$\pm$1.4 \textcolor{dimRed}{(+108.3)}& 8.2$\pm$1.4 \textcolor{dimRed}{(+39.0)} & \textbf{11.4$\pm$1.6} \textcolor{dimGreen}{(-39.7)} & \textbf{1.3$\pm$0.9} \textcolor{dimGreen}{(-84.0)}& 26.7$\pm$7.1 \textcolor{dimRed}{(+36.2)}\\
& MTLR & 4.0$\pm$2.0 \textcolor{dimRed}{(+14.3)} & 2.1$\pm$1.0 \textcolor{dimRed}{(+31.2)}& 2.1$\pm$0.6 \textcolor{dimRed}{(+50.0)} & 7.7$\pm$1.8 \textcolor{dimRed}{(+8.5)} & 1.6$\pm$1.2 \textcolor{dimRed}{(+33.3)}& \textbf{5.1$\pm$1.1} \textcolor{dimGreen}{(-13.6)} & 16.3$\pm$3.4 \textcolor{dimGreen}{(-13.8)} & 4.7$\pm$3.5 \textcolor{dimGreen}{(-42.0)}& 22.4$\pm$9.5 \textcolor{dimRed}{(+14.3)}\\
& BNNSurv & 3.2$\pm$1.7 \textcolor{dimGreen}{(-8.6)} & 1.6$\pm$0.7 \textcolor{dimGreen}{(0.0)}& 1.5$\pm$1.0 \textcolor{dimRed}{(+7.1)} & 6.2$\pm$1.5 \textcolor{dimGreen}{(-12.7)} & 1.4$\pm$0.7 \textcolor{dimRed}{(+16.7)}& 6.8$\pm$1.2 \textcolor{dimRed}{(+15.3)} & 13.2$\pm$3.7 \textcolor{dimGreen}{(-30.2)} & 3.8$\pm$1.3 \textcolor{dimGreen}{(-53.1)}& 28.1$\pm$10.4 \textcolor{dimRed}{(+43.4)}\\
\cmidrule(lr){1-1}
\multirow{5}{*}{\makecell{$\mathcal{D}_{C3}$ (low)\\($N=4930$,\\$d=12$)}}
& CoxPH & 9.8$\pm$2.9 & \textbf{2.7$\pm$1.8} & \textbf{6.1$\pm$2.4} & 26.1$\pm$4.7 & \textbf{3.1$\pm$1.8} & \textbf{27.0$\pm$2.5} & 68.7$\pm$9.6 & 29.4$\pm$10.9& \textbf{64.8$\pm$18.7} \\
& GBSA & 8.7$\pm$2.8 \textcolor{dimGreen}{(-11.2)} & 4.9$\pm$2.5 \textcolor{dimRed}{(+81.5)}& 8.9$\pm$2.4 \textcolor{dimRed}{(+45.9)} & 27.0$\pm$5.7 \textcolor{dimRed}{(+3.4)} & 4.4$\pm$1.6 \textcolor{dimRed}{(+41.9)}& 29.0$\pm$2.6 \textcolor{dimRed}{(+7.4)} & 65.2$\pm$9.9 \textcolor{dimGreen}{(-5.1)} & 24.4$\pm$10.2 \textcolor{dimGreen}{(-17.0)}& 71.8$\pm$17.2 \textcolor{dimRed}{(+10.8)}\\
& RSF & \textbf{6.5$\pm$2.0} \textcolor{dimGreen}{(-33.7)} & 4.8$\pm$2.1 \textcolor{dimRed}{(+77.8)}& 7.8$\pm$2.6 \textcolor{dimRed}{(+27.9)} & \textbf{14.0$\pm$3.0} \textcolor{dimGreen}{(-46.4)} & 12.9$\pm$3.4 \textcolor{dimRed}{(+316.1)}& 35.9$\pm$2.8 \textcolor{dimRed}{(+33.0)} & 39.9$\pm$12.0 \textcolor{dimGreen}{(-41.9)} & \textbf{8.9$\pm$10.4} \textcolor{dimGreen}{(-69.7)}& 94.6$\pm$23.1 \textcolor{dimRed}{(+46.0)}\\
& MTLR & 8.3$\pm$2.7 \textcolor{dimGreen}{(-15.3)} & 3.7$\pm$2.1 \textcolor{dimRed}{(+37.0)}& 7.0$\pm$2.5 \textcolor{dimRed}{(+14.8)} & 19.8$\pm$2.7 \textcolor{dimGreen}{(-24.1)} & 7.9$\pm$3.0 \textcolor{dimRed}{(+154.8)}& 31.2$\pm$2.5 \textcolor{dimRed}{(+15.6)} & 53.9$\pm$9.8 \textcolor{dimGreen}{(-21.5)} & 11.5$\pm$5.3 \textcolor{dimGreen}{(-60.9)}& 76.9$\pm$22.2 \textcolor{dimRed}{(+18.7)}\\
& BNNSurv & 7.4$\pm$3.0 \textcolor{dimGreen}{(-24.5)} & 4.4$\pm$2.5 \textcolor{dimRed}{(+63.0)}& 8.1$\pm$2.7 \textcolor{dimRed}{(+32.8)} & 15.6$\pm$3.6 \textcolor{dimGreen}{(-40.2)} & 14.0$\pm$3.7 \textcolor{dimRed}{(+351.6)}& 39.4$\pm$3.1 \textcolor{dimRed}{(+45.9)} & \textbf{28.1$\pm$5.6} \textcolor{dimGreen}{(-59.1)} & 29.3$\pm$5.9 \textcolor{dimGreen}{(-0.3)}& 125.5$\pm$19.1 \textcolor{dimRed}{(+93.7)}\\
\bottomrule
\end{tabular}
}
\caption{Mean $\pm$ standard deviation of the eMAE, averaged over five folds, with relative improvement/degradation compared to the baseline CoxPH model. Lower values of eMAE indicates that the respective MAE is closer to the true MAE. Generally, MAE\textsubscript{M} provides a good estimate of the true MAE in these datasets. $N$: sample size, $d$: number of features.}
\label{tab:xjtu_emae}
\end{table*}

\begin{table}[!htbp]
\centering
\resizebox{\columnwidth}{!}{%
\begin{tabular}{@{} cc llll @{}}
\toprule
\multirow{2}{*}{\makecell{Dataset}} &
\multirow{2}{*}{\makecell{Model}} &
\multicolumn{3}{c}{\makecell{True MAE}} \\
\cmidrule(lr){3-5}
& & $C=25\%$ & $C=50\%$ & $C=75\%$ \\
\midrule
\multirow{5}{*}{\makecell{$\mathcal{D}_{C1}$\\(high)\\($N=635$,\\$d=12$)}}
& LASSO$^\ast$ & 18.5$\pm$1.1 & 19.6$\pm$0.8 & 19.7$\pm$1.3\\
& CoxPH & 14.7$\pm$1.1 \textcolor{dimGreen}{(-20.5)} & 15.5$\pm$1.5 \textcolor{dimGreen}{(-20.9)} & 17.3$\pm$0.7 \textcolor{dimGreen}{(-12.2)}\\
& GBSA & 17.4$\pm$1.5 \textcolor{dimGreen}{(-5.9)} & 19.5$\pm$1.5 \textcolor{dimGreen}{(-0.5)} & 23.0$\pm$1.5 \textcolor{dimRed}{(+16.8)}\\
& RSF & \textbf{12.6$\pm$0.8} \textcolor{dimGreen}{(-31.9)} & \textbf{14.0$\pm$1.7} \textcolor{dimGreen}{(-28.6)} & \textbf{17.2$\pm$1.2} \textcolor{dimGreen}{(-12.7)}\\
& MTLR & 13.8$\pm$1.1 \textcolor{dimGreen}{(-25.4)} & 14.6$\pm$2.4 \textcolor{dimGreen}{(-25.5)} & 17.5$\pm$2.4 \textcolor{dimGreen}{(-11.2)}\\
& BNNSurv & 14.2$\pm$1.1 \textcolor{dimGreen}{(-23.2)} & 15.3$\pm$2.1 \textcolor{dimGreen}{(-21.9)} & 17.8$\pm$1.7 \textcolor{dimGreen}{(-9.6)}\\
\cmidrule(lr){1-1}
\multirow{5}{*}{\makecell{$\mathcal{D}_{C2}$\\(medium)\\($N=1586$,\\$d=12$)}}
& LASSO$^\ast$ & 63.0$\pm$3.6 & 64.8$\pm$4.0 & 65.0$\pm$3.2\\
& CoxPH & 56.4$\pm$3.8 \textcolor{dimGreen}{(-10.5)} & 57.1$\pm$2.9 \textcolor{dimGreen}{(-11.9)} & 70.3$\pm$14.6 \textcolor{dimRed}{(+8.2)}\\
& GBSA & 60.8$\pm$3.0 \textcolor{dimGreen}{(-3.5)} & 67.0$\pm$2.0 \textcolor{dimRed}{(+3.4)} & 81.9$\pm$6.3 \textcolor{dimRed}{(+26.0)}\\
& RSF & \textbf{31.5$\pm$2.0} \textcolor{dimGreen}{(-50.0)} & \textbf{34.3$\pm$4.5} \textcolor{dimGreen}{(-47.1)} & \textbf{43.2$\pm$7.0} \textcolor{dimGreen}{(-33.5)}\\
& MTLR & 50.9$\pm$4.1 \textcolor{dimGreen}{(-19.2)} & 53.1$\pm$3.9 \textcolor{dimGreen}{(-18.1)} & 59.1$\pm$5.3 \textcolor{dimGreen}{(-9.1)}\\
& BNNSurv & 49.2$\pm$4.3 \textcolor{dimGreen}{(-21.9)} & 49.7$\pm$2.1 \textcolor{dimGreen}{(-23.3)} & 53.2$\pm$3.6 \textcolor{dimGreen}{(-18.0)}\\
\cmidrule(lr){1-1}
\multirow{5}{*}{\makecell{$\mathcal{D}_{C3}$\\(low)\\($N=4930$,\\$d=12$)}}
& LASSO$^\ast$ & 206.4$\pm$5.1 & 209.9$\pm$8.4 & 224.0$\pm$6.1\\
& CoxPH & 183.7$\pm$6.2 \textcolor{dimGreen}{(-11.0)} & 190.6$\pm$5.9 \textcolor{dimGreen}{(-9.2)} & 265.5$\pm$34.2 \textcolor{dimRed}{(+18.5)}\\
& GBSA & 197.4$\pm$2.6 \textcolor{dimGreen}{(-4.4)} & 216.2$\pm$4.2 \textcolor{dimRed}{(+3.0)} & 270.8$\pm$5.2 \textcolor{dimRed}{(+20.9)}\\
& RSF & \textbf{105.3$\pm$6.2} \textcolor{dimGreen}{(-49.0)} & \textbf{110.7$\pm$7.5} \textcolor{dimGreen}{(-47.3)} & \textbf{136.2$\pm$1.9} \textcolor{dimGreen}{(-39.2)}\\
& MTLR & 151.7$\pm$6.6 \textcolor{dimGreen}{(-26.5)} & 159.0$\pm$8.2 \textcolor{dimGreen}{(-24.2)} & 191.0$\pm$6.4 \textcolor{dimGreen}{(-14.7)}\\
& BNNSurv & 139.8$\pm$5.0 \textcolor{dimGreen}{(-32.3)} & 141.0$\pm$8.5 \textcolor{dimGreen}{(-32.8)} & 149.5$\pm$4.6 \textcolor{dimGreen}{(-33.3)}\\
\bottomrule
\end{tabular}%
}
\caption{Mean $\pm$ standard deviation of the true MAE, averaged over five folds, with relative improvement/degradation compared to the baseline LASSO model. $^\ast$The LASSO model is a linear model trained on the uncensored observations only, thus discarding the censored ones. In most cases, using a model that supports censored data gives a notable improvement in predictive accuracy over the LASSO baseline. $N$: sample size, $d$: number of features, $C$: the level of censoring.}
\label{tab:xjtu_true_mae}
\end{table}

\begin{table}[!hbp]
\centering
\resizebox{\columnwidth}{!}{%
\begin{tabular}{cccccccc}
\toprule
\multirow{2}{*}{\makecell{Dataset}} &
\multirow{2}{*}{\makecell{Model}} &
\multicolumn{2}{c}{25\% censoring} & \multicolumn{2}{c}{50\% censoring} & \multicolumn{2}{c}{75\% censoring} \\
\cmidrule(lr){3-4}\cmidrule(lr){5-6}\cmidrule(lr){7-8}
& & D-Cal & C-Cal & D-Cal & C-Cal & D-Cal & C-Cal \\
\midrule
\multirow{5}{*}{\makecell{$\mathcal{D}_{C1}$ (high)\\($N=635$, $d=12$)}}
& CoxPH & 5/5 & - & 5/5 & - & 5/5 & - \\
& GBSA & 5/5 & - & 5/5 & - & 5/5 & - \\
& RSF & 5/5 & - & 5/5 & - & 5/5 & - \\
& MTLR & 5/5 & - & 5/5 & - & 5/5 & - \\
& BNNSurv & 3/5 & 5/5 & 5/5 & 0/5 & 4/5 & 0/5 \\
\cmidrule(lr){1-1}
\multirow{5}{*}{\makecell{$\mathcal{D}_{C2}$ (medium)\\($N=1586$, $d=12$)}}
& CoxPH & 5/5 & - & 5/5 & - & 5/5 & - \\
& GBSA & 4/5 & - & 5/5 & - & 5/5 & - \\
& RSF & 4/5 & - & 5/5 & - & 5/5 & - \\
& MTLR & 5/5 & - & 5/5 & - & 5/5 & - \\
& BNNSurv & 4/5 & 5/5 & 5/5 & 2/5 & 5/5 & 0/5 \\
\cmidrule(lr){1-1}
\multirow{5}{*}{\makecell{$\mathcal{D}_{C3}$ (low)\\($N=4930$, $d=12$)}}
& CoxPH & 3/5 & - & 5/5 & - & 5/5 & - \\
& GBSA & 5/5 & - & 5/5 & - & 5/5 & - \\
& RSF & 0/5 & - & 3/5 & - & 5/5 & - \\
& MTLR & 5/5 & - & 5/5 & - & 5/5 & - \\
& BNNSurv & 1/5 & 5/5 & 1/5 & 0/5 & 1/5 & 0/5 \\
\bottomrule
\end{tabular}
}
\caption{Number of times the mean survival function and the individual survival distributions were D-calibrated and C-calibrated, respectively, according to a goodness-of-fit test over five folds. A model is calibrated if it has a $p$-value greater than 0.05. C-calibration is only applicable for models that can predict credible intervals (CrIs)~\citep{qi_using_2023}. MTLR is the only model that gives D-calibrated survival curves on all datasets, no matter the level of censoring. $N$: sample size, $d$: number of features.}
\label{tab:xjtu_calibration}
\end{table}

\subsection{Stratified survival curves}
\label{sec:stratified_survival_curves}

\begin{figure*}[!htbp]
     \centering
     \begin{subfigure}[b]{0.9\textwidth}
         \centering
         \includegraphics[width=1\textwidth]{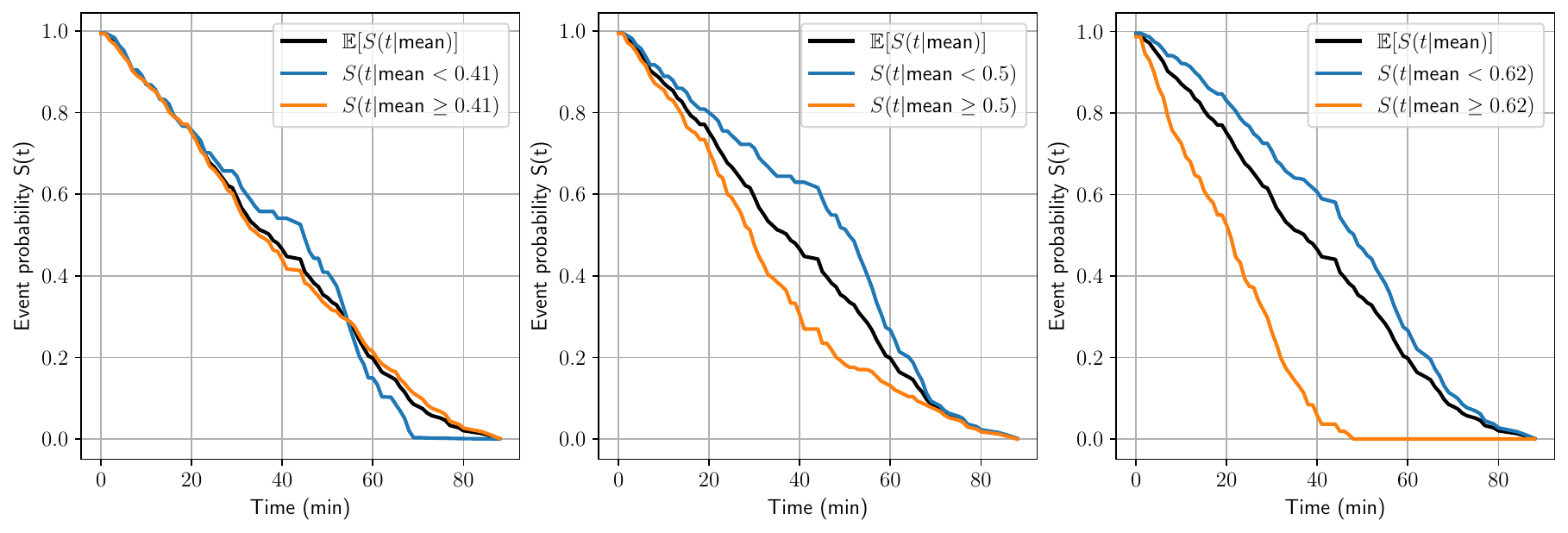}
         \caption{The ``mean'' feature stratified by the 25th, 50th and 75th percentiles.}
         \label{fig:group_survival_mean}
     \end{subfigure}
     \hfill
     \begin{subfigure}[b]{0.9\textwidth}
         \centering
         \includegraphics[width=1\textwidth]{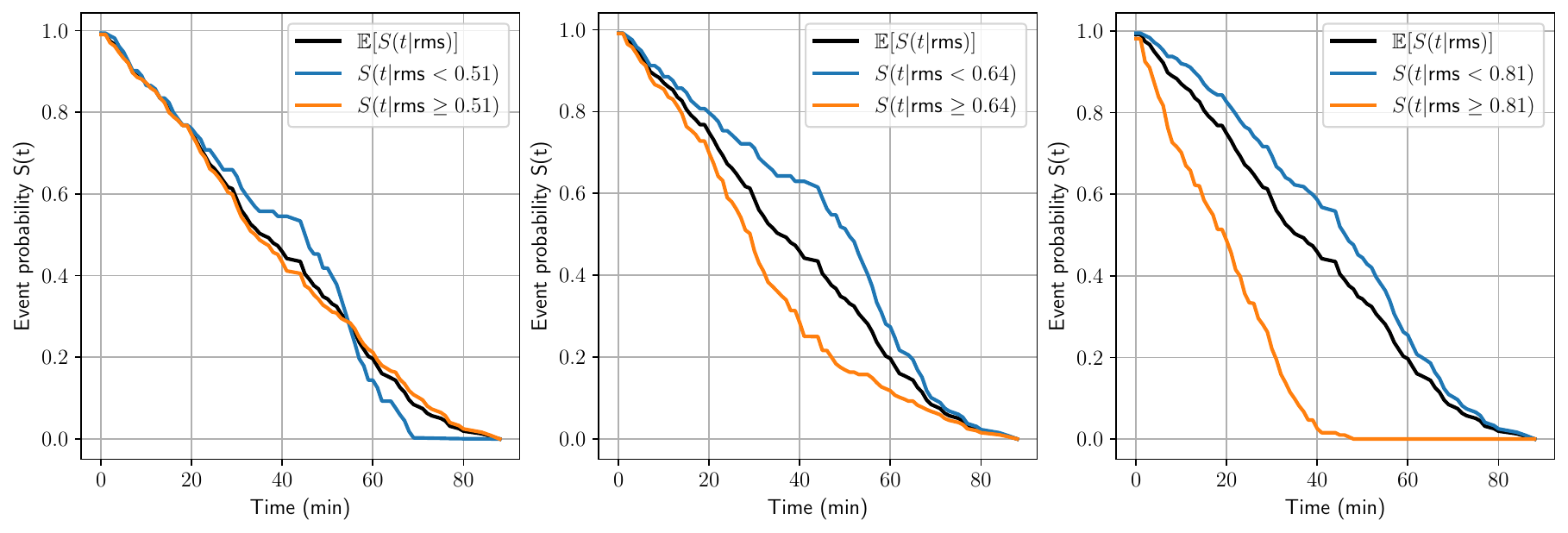}
         \caption{The ``rms'' feature stratified by the 25th, 50th and 75th percentiles.}
         \label{fig:group_survival_rms}
     \end{subfigure}
     \hfill
     \begin{subfigure}[b]{0.9\textwidth}
         \centering
         \includegraphics[width=1\textwidth]{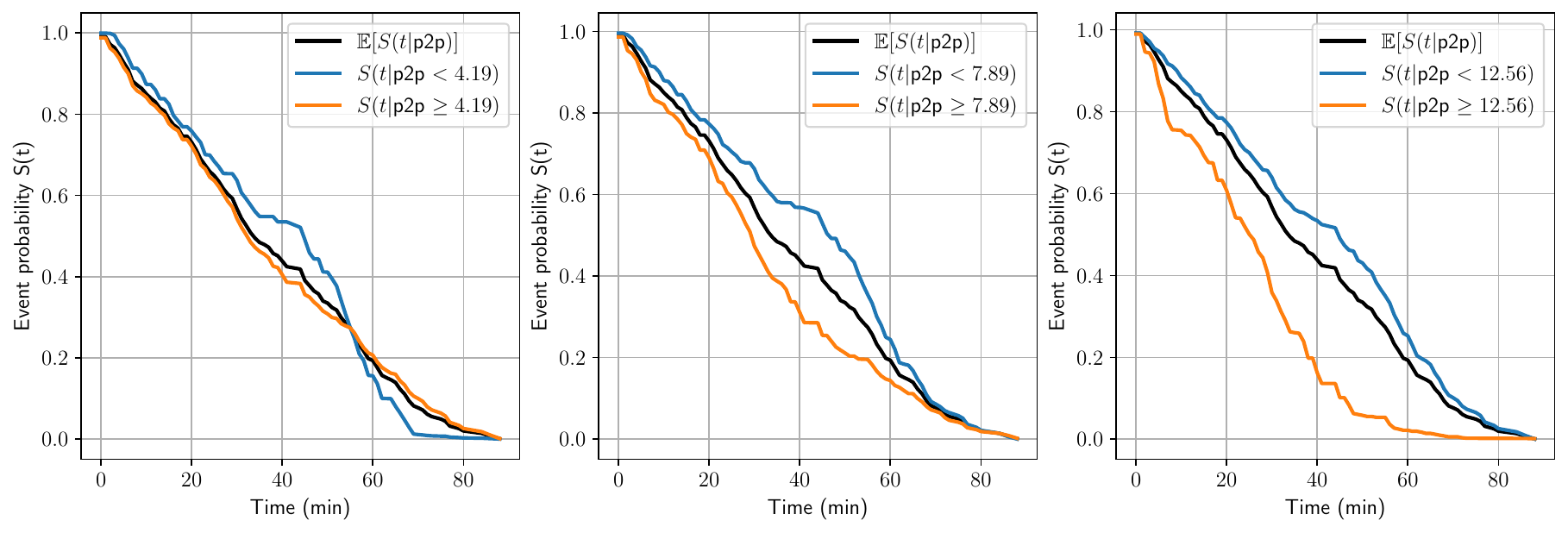}
         \caption{The ``p2p'' feature stratified by the 25th, 50th and 75th percentiles.}
         \label{fig:group_survival_p2p}
     \end{subfigure}
        \caption{Predicted survival curves over a stratified set of features using the RSF model~\citep{ishwaran_random_2008} (Black line: the mean predicted survival function without stratification; Blue line: the predicted survival function for bearings with features below the respective percentile; Orange line: the predicted survival function for bearings with features above). We set $C=25\%$ and train the model on the $\mathcal{D}_{C1}$ dataset, where 80\% of the data is used for training and 20\% is used for testing. The curves are plotted only on the test set.}
        \label{fig:stratified_survival_curves}
\end{figure*}

Figure \ref{fig:stratified_survival_curves} shows the predicted mean survival curve and two curves where the features have been stratified according to their 25th, 50th and 75th percentiles from left to right. The $\mathcal{D}_{C1}$ dataset is used for this experiment with 25\% censoring. Recall that the survival function, $S(t)$, is the probability that the event occurs later than time $t$, so higher values are associated with a lower predicted risk of the event (failure). We see that the predicted survival curves have notable differences based on the feature's value and that some feature values lead to significantly lower survival probabilities. In fact, across all features evaluated in this plot, the RSF model produces more favorable survival outcomes when the value is below the median and unfavorable outcomes when it is above (center column in Figure \ref{fig:stratified_survival_curves}). For example, the root mean square (RMS) value represents the amount of energy in a signal. As the bearing begins to deteriorate, the RMS value will increase (center row in Figure \ref{fig:stratified_survival_curves}). This happens because the number of peaks increases, which increases the total energy of the signal.

Typically, during the initial phases of a mechanical fault, the RMS value exhibits minimal alteration due to the limited shift in the total energy of the signal. However, as the bearing begins to deteriorate, the RMS value tends to show a more pronounced increase. This way of stratification can provide insight into the learned representation between the inputs and the outputs, and can create counterfactual explanations: If the signal energy was less, the bearing would have failed later.

\subsection{Individual RUL prediction}
\label{sec:individual_rul_prediction}

\begin{figure*}[!htbp]
    \centering
    \includegraphics[width=0.9\linewidth]{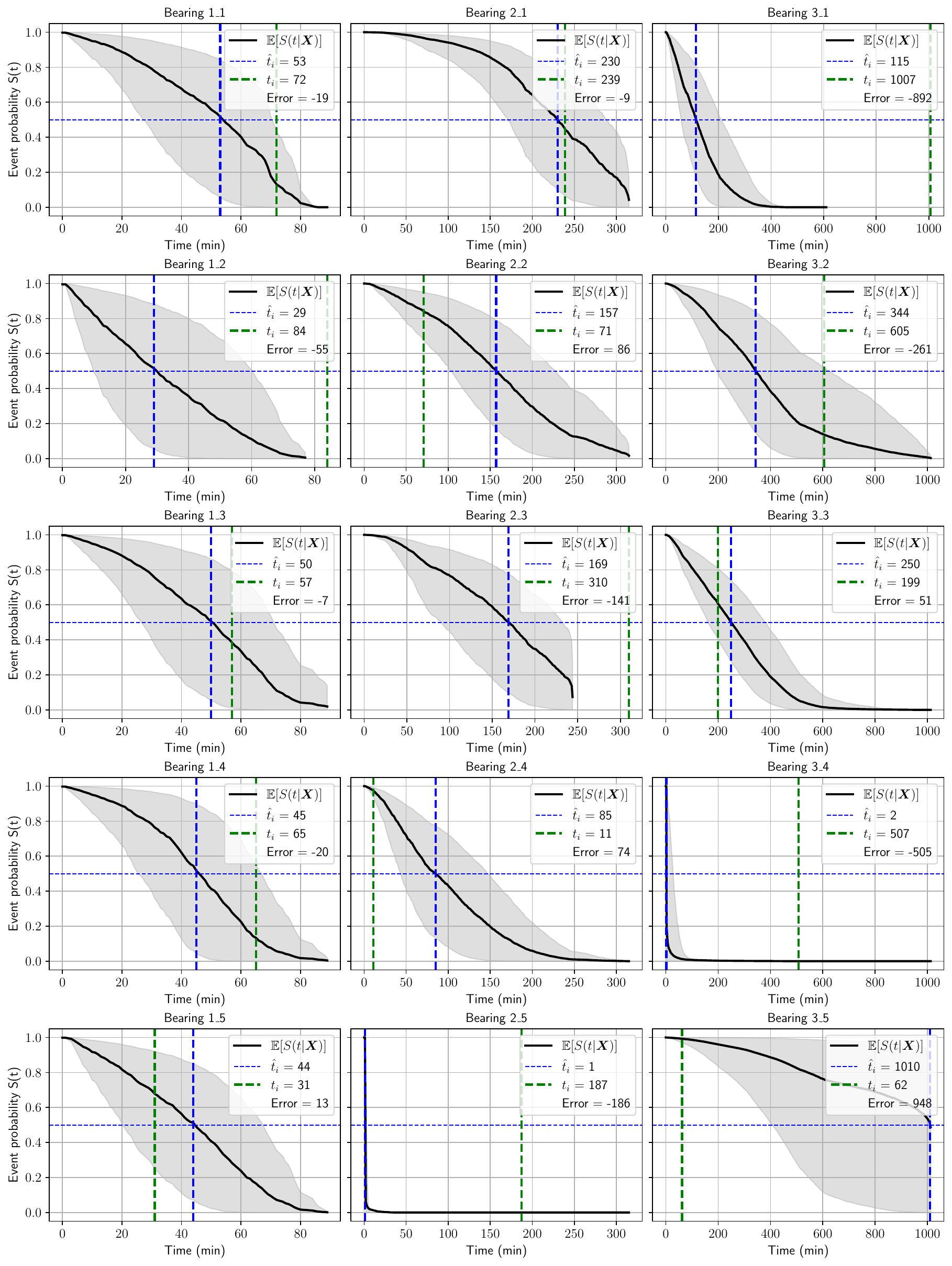}
    \caption{Predicted individual survival distribution (ISD) for each bearing using the BNNSurv model~\citep{lillelund_uncertainty_2023} (Blue line: predicted event time $\hat{t}_{i}$; Green line: actual event time $t_i$; Shaded grey: the region between the upper and lower bounds of a 90\% credible interval). Bearings are separated by their operating condition (three columns) and index (five rows). The prediction error is the difference between the predicted time (the cross point of the vertical and horizontal blue lines) and the observed event time (green dash-dotted line).}
    \label{fig:isd}
\end{figure*}

We use the BNNSurv model to make individual RUL predictions per bearing. This is also called the individual survival distribution (ISD). In this case, the ISD is a representation of the survival probability for a single bearing at different time points based on its current features. For each operating condition, we have a total of five bearings, when sampling from the $X$-axis, as suggested by~\cite{wang_hybrid_2020}. To predict the ISD of a single bearing, we train $M$ models, where $M$ is the number of operating conditions multiplied by the number of bearings. For example, to predict the ISD of the bearing ($i,j$), where $i$ is the operating condition and $j$ is the bearing index, we train the model $M_{i,j}$ in the dataset $\mathcal{D}_{i}^{-j}$, where the notation $-j$ denotes that the dataset excludes the $j$th bearing. We then predict the ISD by taking a single sample from bearing ($i,j$) at the beginning of its lifetime. We repeat this process iteratively to estimate the ISD for all bearings. Figure \ref{fig:isd} shows the predicted ISDs and the upper and lower bounds of the credible intervals (CrIs), estimated by drawing 100 samples from the predictive distribution. We follow~\cite{qi_using_2023} and take the median of the survival curve (0.5) as the predicted time-to-event, $\hat{t}_{i}$, and compare it with the observed event time, $t_{i}$. Although the event detector always detects the event of interest before the end of life (see Table \ref{tab:xjtu_ed_actual}), the RUL can be predicted before or after this point. We see a lot of uncertainty in many of the predictions (wide credible intervals), especially for the C1 operating condition due to the low amount of training data available (left column), and there are many irregularities in the predicted RUL, especially for conditions C2 and C2 (center and right columns), which the model cannot reliably predict from just a single sample in this case. Ideally, the predicted time-to-event should match the observed time-to-event and preferably before the end of life, so that predictive maintenance can be initiated in time. Depending on the industrial application and the costs of predictive maintenance, overestimating RUL is often worse than underestimating it.

\subsection{Comparison to SOTA}

To compare our approach to literature methods, we adopt the cumulative relative accuracy (CRA) metric~\citep{saxena_metrics_2021}. This metric can fully assess the accuracy of a prognostic approach by aggregating the relative prediction accuracies at any inspection time. The CRA is calculated as follows:

\begin{equation}\label{eq:cra}
CRA = K \sum_{k=1}^{K} w_k \, RA(T_k),
\end{equation}

\noindent where $K$ is the number of inspection windows, $w_{k}$ is a normalized weight factor with $\frac{k}{\sum_{k=1}^{K} k}$ and $RA(T_k)$ is the relative prediction accuracy at inspection time $T_k$:

\begin{equation}\label{ratk}
RA(T_k) = 1 - \frac{\left| \text{ActRUL}(T_k) - \text{RUL}(T_k) \right|}{\text{ActRUL}(T_k)},
\end{equation}

\noindent where ActRUL($T_k$) is the actual RUL at inspection time $T_k$, and RUL($T_k$) is the estimated RUL. The closer the CRA to one, the more accurate the prediction.

Table \ref{tab:sota_comparasion} presents the performance evaluation results of the six commonly-used RUL prediction methods and the proposed method on the XJTU-SY dataset. Since none of the methods studied provided their source code, it was not possible to reproduce the results, so the CRA is reported as in~\cite{wang_hybrid_2020}. For the proposed method, we set the inspection time $T_k=1$, \ie at the beginning of the recording, and take a single sample the $X$-axis. We then train a Random Survival Forests model based on the features and time-to-event information of the remaining four bearings and make the RUL prediction on the holdout bearing, \ie training a model for each individual bearing iteratively.

The results show that our method achieves state-of-the-art predictive performance under operating condition C1. We attribute this to the relatively short event horizon of the bearings in this category. By the time of inspection, the bearings are already nearing failure, as evidenced by the sensor readings. As a result, the data collected at this stage are highly indicative of the impending event. Under a decreased load with a longer event horizon, \ie conditions C2 and C3, sampling data from the beginning of the recording is no longer indicative of when the bearing is going to fail. We see that the predictive errors fluctuate a lot and the proposed method generally performs worse than the Hybrid model~\citep{wang_hybrid_2020} under conditions C2 and C3. We attribute this behavior to the variance in the XJTU-SY dataset, as some bearings have entirely different lifespans, and also to the lack of good predictive features at inspection time. At $T_k=1$, when the bearing is still working properly, the prediction might be less reliable since there is little or no evidence of wear or damage at that early point. On the other hand, sampling from the bearing at a later point in time than $T_k=1$ may introduce survivorship bias in this case: If we only make predictions after a certain amount of time has passed (\eg ten minutes), we exclude bearings that might have failed earlier, leading to survivorship bias. This can skew the data toward more durable bearings and give an overly optimistic view of how long a specific bearing would last, since we are assuming that the bearing survives past a certain minute mark. In this particular example, we also have fewer training samples at our disposal than the literature methods, since we estimate the RUL as the time to failure rather than the end of life. In some cases, this means that we only have a quarter of the training data available compared to the other methods (see Table \ref{tab:xjtu_ed_actual}).

We would have liked to compare our method with~\cite{WANG2023109747}, who also used the XJTU-SY dataset and proposed a compelling two-stage prediction method; however, the authors only provided aggregated results of the CRA. Similarly,~\cite{xu_novel_2023} proposed a multiscale encoder-decoder consisting of a CNN and a multi-head attention mechanism, but the authors only provided a single RMS error, not an individual error for each bearing.~\cite{SU2021107531} proposed a deep adaptive transformer model, which supposedly provides better forecasting performance and solves the problem of vanishing gradients in existing RNN architectures, but the reported relative accuracy is only for 2 of the 15 bearings in the XJTU-SY dataset. In addition, the authors did not provide source code to reproduce their results.

\begin{table*}[!htbp]
\centering
\resizebox{0.7\textwidth}{!}{
\begin{tabular}{ccccccccc}
\toprule
{\centering\shortstack{{Operating}\\{condition}}} &
{\centering\shortstack{{Bearing}\\{dataset}}} &
RVM (SL) & DBN (DL) & PF (SL) & EKF (SL) & Hybrid (SL) & {\centering\shortstack{RULSurv (SL)}} \\
\midrule
\multirow{5}{*}{\shortstack{C1\\(high)}}
& Bearing 1\_1 & 0.5741 & 0.4318 & 0.6107 & 0.6209 & \textbf{0.9047} & 0.8745 \\
& Bearing 1\_2 & 0.1815 & 0.6248 & 0.7256 & 0.3500 & \textbf{0.8546} & 0.5501 \\
& Bearing 1\_3 & 0.6245 & 0.5571 & 0.4850 & 0.8010 & 0.8482 & \textbf{0.8782} \\
& Bearing 1\_4 & 0.3722 & -0.9479 & 0.2305 & 0.6839 & \textbf{0.7240} & 0.6922 \\
& Bearing 1\_5 & 0.6122 & 0.6636 & 0.4311 & 0.5042 & 0.7878 & \textbf{0.7980} \\
\cmidrule(lr){1-1}
\multirow{5}{*}{\shortstack{C2\\(medium)}}
& Bearing 2\_1 & 0.5718 & 0.5518 & 0.3963 & 0.5150 & \textbf{0.8621} & 0.7649 \\
& Bearing 2\_2 & 0.1789 & -0.1977 & 0.2634 & 0.4314 & \textbf{0.6521} & -0.7065 \\
& Bearing 2\_3 & 0.6172 & 0.9013 & 0.7364 & 0.8800 & \textbf{0.9612} & 0.4041 \\
& Bearing 2\_4 & -0.0693 & 0.5316 & 0.4633 & 0.5004 & \textbf{0.6276} & -8.3418 \\
& Bearing 2\_5 & 0.2563 & 0.0671 & 0.1833 & 0.4815 & \textbf{0.6328} & 0.0388  \\
\cmidrule(lr){1-1}
\multirow{5}{*}{\shortstack{C3\\(low)}}
& Bearing 3\_1 & 0.4329 & 0.6979 & 0.6557 & 0.7744 & \textbf{0.8942} & 0.1936 \\
& Bearing 3\_2 & 0.2225 & -0.1301 & 0.1518 & 0.5362 & \textbf{0.6517} & 0.3039 \\
& Bearing 3\_3 & -0.0883 & 0.5167 & 0.1283 & -0.9653 & \textbf{0.8887} & -0.5153 \\
& Bearing 3\_4 & 0.6570 & 0.6050 & 0.7830 & 0.6670 & \textbf{0.8133} & 0.6112 \\
& Bearing 3\_5 & 0.4881 & 0.5618 & 0.3857 & 0.5040 & \textbf{0.6512} & -1.4658 \\
\bottomrule
\end{tabular}
}
\caption{Cumulative relative accuracy (CRA) results of six RUL prediction models. This comparison includes Relevance Vector Machines (RVM)~\citep{DIMAIO2012405}, Deep Belief Network (DBN)~\citep{7508982}, Particle Filtering (PF)~\citep{6544227}, Extended Kalman Filtering (EKF)~\citep{6850072}, the Hybrid model~\citep{wang_hybrid_2020}, and our method. ``SL'' denotes a shallow-learning architecture, ``DL'' denotes a deep-learning architecture.}
\label{tab:sota_comparasion}
\end{table*}

\section{Discussion}

\subsection{Limitations} \label{sec:limitations}

We propose an event detection algorithm based on the KL divergence. This offers interpretability, as opposed to using a DL architecture, but also requires manual feature engineering and fine-tuning, since we have to carefully select the frequency bands and calibrate the threshold function to avoid false positives and false negatives. In addition, there are some challenges in using the proposed frequency spectra to identify anomalies.

\begin{enumerate}[1)]
\item The fault frequency is based on the assumption that no sliding occurs between the rolling element and the bearing raceway, \ie these rolling elements will only roll on the raceway. Under low load, this is rarely the case, as the rolling element often undergoes a combination of rolling and sliding movement if the load is insufficient to overcome the friction imposed by the lubricant. As a consequence, the calculated frequency may deviate from the real fault frequency and make this manually determined feature less informative of a bearing defect.
\item In all rotating machinery, the observed vibration signature will be a sum of contributions from mechanical components in the system; bearing defects, bearing looseness, misalignment of shafts from \eg the motor to the gearbox, imbalances in the rotating parts, or gear-teeth meshing. These contributions will interact and the resultant characteristic frequencies can be added or subtracted with a magnitude determined by the sensor location and how mobile each component is, obfuscating the informative frequencies.
\item Bearing preload and clearance will affect the eigenfrequencies of the rolling elements and as a result lead to changes in magnitude and a shift in the frequency bands that are excited during the overroling of infant bearing defects. Furthermore, changes in temperature can also lead to thermal expansion that affects the contact angle between the roller and the raceway, leading to volatile bearing frequencies.
\item Additional factors, such as the quality of lubrication, the roughness of the surface of the rolling element, and changes in surface conformance, do not manifest themselves as characteristic cyclic frequencies, making them difficult to detect with traditional model-based spectral analysis or classical data-driven ML methods.
\item The sensitivity of various features that indicate a defect may vary considerably under different operating conditions. A very thorough and systematic learning stage is typically required to test the sensitivity of these frequencies under any desirable operating condition before they can actually be put into use with a traditional approach.
 \end{enumerate}

Our method supports several survival models (CoxPH, GBSA, and BNNSurv) that assume proportional hazards, \ie the relative risk of an event is independent of time. This assumption is easier to satisfy when the event horizon is short. In an industrial setting, where bearings usually last longer than under artificially accelerated degradation, the time-to-event distribution may have temporal dependencies at different times that may not be easily captured by a proportional hazard model.

Based on the estimated survival curve, we use the median of the curve (0.5) as the predicted time-to-event. Other approaches have been proposed, such as using the area under the survival function (AUC) as the predicted time, or simulating the survival times based on some probability distribution~\citep{austin_generating_2012}. Although it is unlikely that a single bearing will fail at the exact time the survival curve reaches 0.5, a population of bearings will fail at that time on average, given a sufficiently large sample size and that the survival curves are D-calibrated. The XJTU-SY dataset includes only five bearings per operating condition, and sensor readings show considerable variability even among bearings tested under the same load configuration. This variability contributes to inaccuracies in predicting the exact RUL.

\subsection{Industrial applicability} \label{sec:industrial_applicability}

The proposed event detection algorithm showed strong performance in the XJTU-SY data set, since it always predicted the event of interest before the end of life, facilitating timely predictive maintenance. However, extending the event detector to a different dataset, \eg PRONOSTIA~\citep{nectoux2012pronostia}, IMS~\citep{lee2007bearing}, or an industrial application, would require fine-tuning of the algorithm. In particular, the parameter $\lambda_{KL}$, which controls the algorithm's sensitivity as it approaches the end of the bearing's life. The proposed algorithm also assumes that the end of life is known in advance to calibrate the event detection part. In a real-world scenario, this information may not be available, so we recommend using $L_{10}$ instead in such situations.

Moreover, the effectiveness of the proposed method is based on the assumption that defects progress in a predictable and gradual manner, starting from the initial stages and advancing through a series of deterioration steps (see Fig. \ref{fig:defect_preogression}). This steady development of bearing defects, whether they occur alone or in combination, would facilitate detailed monitoring and data collection in industrial settings, due to the gradual changes in vibration patterns. However, despite this gradual progression, the variation in vibration signatures among even similar rotating machinery can be significant, as discussed in Section \ref{sec:limitations}. In contrast, the vibration patterns observed in the XJTU-SY dataset are the result of excessive load on the bearings, preventing proper lubrication and leading to immediate surface damage and rapid degradation of the raceway. Although the proposed method performed well on the XJTU-SY dataset, which features bearings that undergo accelerated degradation, industrial settings often present a different challenge, where degradation is less apparent and wear develops gradually over time. Given this disparity, practitioners need to carefully select the machine-specific thresholds for the event detector and have a sufficient hold-out of set of multiple bearings to evaluate the performance of the predictive model. Whenever the model is used for a novel application, the event detector and the predictive model will have to be reevaluated. In summary, the extension to other types of data sets and industrial applications poses the following challenges:

\begin{itemize}
\item Industrial sensor data are often noisier and less reliable than controlled experimental data. In an industrial setting, sensor failures, environmental factors, and other variables could introduce noise into the training dataset. In the presented work, we do not perform any outlier detection before training the models. However, if significant noise or anomalies are present in the input data, incorporating an anomaly detection algorithm may be necessary.
\item Adapting the event detector to a new dataset requires fine-tuning the algorithm, particularly the parameter $\lambda_{KL}$, which controls the algorithm's sensitivity. Moreover, in a real industrial application, we recommend using the $L_{10}$ instead of the end of life to calibrate the algorithm, as the latter may not be available.
\item The survival models are only evaluated in the XJTU-SY dataset, where bearings are subject to accelerated degradation. In a new dataset, models will have been retrained and reevaluated in a hold-out set. However, using either the MTLR or the BNNSurv model, the learned parameters (weights and biases) can be used as initial weights when training on a new dataset (\ie pretraining).
\end{itemize}

\subsection{Scalability} \label{sec:scaleability}

In this study, we have proposed several survival models for RUL prediction, both neural networks (MTLR and BNNSurv) and non-neural networks (CoxPH, GBSA, and RSF). In terms of model scalability to larger datasets with more observations and a larger number of features, the scalability of the non-neural networks is limited by memory constraints. As datasets grow larger, these methods become less practical as they have to process the entire dataset for each gradient update during training. However, a neural network only loads small batches of the entire training set into memory during training, allowing the system to handle larger datasets efficiently without requiring a proportional increase in memory. This makes MTLR and BNNSurv suitable for training on large-scale datasets that exceed the available memory on a given system. Moreover, using minibatches also allows for better hardware utilization, such as GPUs, as they can process multiple samples in parallel.

\subsection{Future work}\label{sec:future_work}

We find it appealing to test the proposed method on an industrial dataset to validate its practicality. The XJTU-SY dataset only comes with five bearings per operating condition and they have all been subjected to accelerated degradation, which substantially shortens their lifespan compared to actual industrial bearings. A real-world dataset would not have these characteristics. We also want to extend our method to support time-varying features, thus incorporating temporal information between the vibration signals in the model, similar to~\cite{zhang_time-dependent_2019}, however, using our definition of event. This can be done by replacing the learned representation with a recurrent neural network architecture, such as a standard RNN or its variants, \eg GRU, LSTM or Transformer-based architectures. Three of the five proposed models assume proportional hazards; however, the proportional hazards assumption may not hold in an industrial setting. Thus, models that support time-varying features or models that do not rely on the proportional hazards assumption may be more suited for industrial application. Lastly, as a future work, we have considered a few potential alternatives to alleviate the problem of the identification and recalibration of the frequency bands related to the event detector. One promising candidate for alleviating the limitations mentioned above is the Multivariate Variational Decomposition (MVMD) model~\citep{Rehman2019}, which allows decomposing the recordings into their intrinsic oscillatory components, obtaining the different frequency bands in a data-driven way.

\section{Conclusion}

We have introduced a novel method for event detection and RUL estimation in ball bearings that naturally supports censored data. The key outcomes of the proposed method are as follows: (1) It offers an interpretable event detection algorithm based on Kullback-Leibler (KL) divergence to annotate a ball bearing dataset with the time a bearing starts to fail, before the end of life. (2) It provides a flexible set of survival models to predict the RUL, which is guaranteed to decrease monotonically as a function of time. (3) It uses the median of the survival curve as the predicted time-to-event, providing a quantitative and probabilistic estimate of the event occurrence. (4) It gives state-of-the-art predictive performance on the XJTU-SY dataset under the highest load condition. (5) It can make use of uncertainty-aware survival methods, \eg Bayesian methods, which can predict an individual survival distribution with credible intervals for each bearing. The findings of this work have significant implications for real-life predictive maintenance and reliability engineering: (1) Given a ball bearing dataset with fully-observed recordings, the proposed event detector can label each bearing with its time to failure (\ie where the bearing starts to malfunction), not its end of life. Thus, any machine learning model trained on this dataset will learn to predict the time to failure instead of the end of life. (2) By leveraging censored data in predictive modeling, the time-to-event estimates are evidently more accurate than if we were to use only data from bearings with a known failure time. (3) The proposed method includes five survival models, suitable for modeling under different conditions \eg under nonproportional hazards or in nonlinear risk applications. In general, these three contributions can lead to better RUL predictions, which in turn can help industries such as manufacturing, transportation, and energy reduce downtime, lower maintenance costs, and improve operational efficiency.

\section*{CRediT authorship contribution statement}

\textbf{Christian Marius Lillelund:} Conceptualization, Methodology, Writing - Original draft preparation, Software, Validation, Revision. \textbf{Fernando Pannullo:} Conceptualization, Methodology, Writing - Reviewing and Editing, Software, Revision. \textbf{Morten Opprud Jakobsen:} Data Curation, Conceptualization, Writing - Reviewing and Editing, Revision. \textbf{Manuel Morante:} Conceptualization, Methodology, Writing - Reviewing and Editing, Revision. \textbf{Christian Fischer Pedersen:} Conceptualization, Methodology, Supervision, Revision.

\section*{Declaration of competing interest}

The authors declare that they have no known competing financial interests or personal relationships that could have appeared to influence the work reported in this paper.

\section*{Data availability}

The XJTU-SY dataset is freely available. A download link can be found in the source code repository.

\appendix

\section{Evaluation metrics}
\label{app:evaluation_metrics}

\textbf{MAE:} The mean absolute error is the absolute difference between the actual and predicted survival times (defined as the median time of the predicted survival curve). Given an individual survival distribution (ISD), $S\br{t\;|\;\bm{x}_{i}} = \P\br{T>t\;|\;\bm{X} = \bm{x}_{i}}$, we compute the predicted survival time $\hat{t}_i$ as the median survival time~\citep{qi2023effective}:
\begin{equation}\label{eq:yi}
\hat{t}_i = \text{median}\;(S(t\;|\;\bm{x}_{i})) = S^{-1}(\tau = 0.5\;|\;\bm{x}_{i})\text{,}
\end{equation}

\noindent where $\tau \in [0,1]$ represents the quantile probability level. Given $\hat{t}_{i}$ and ${t}_{i}$, it is then trivial to calculate the MAE if the event is observed ($\delta_{i} = 1$)~\citep{qi2023effective}:
\begin{equation}
\text{MAE}\;(\hat{t}_{i}, t_{i}, \delta_{i} = 1) = |\;t_{i}-\hat{t}_{i}\;|\text{.}
\end{equation}

\textbf{MAE\textsubscript{H}:} In cases where the event is not observed ($\delta_{i} = 0$), there are still ways to calculate the MAE. One approach is using the hinge loss function for the censored observations: if the predicted survival time $\hat{t_{i}}$ is less than the censoring time, $c_{i}$, the loss equals the censoring time minus the predicted time. When the predicted survival time is equal to or exceeds the censoring time, the loss is zero~\citep{qi2023effective}:
\begin{equation}
\text{MAE-Hinge}\;(\hat{t}_{i}, t_{i}, \delta_{i} = 0) = \text{max}\;\br{t_{i}-\hat{t}_{i},\;0}\text{.}
\end{equation}

\textbf{MAE\textsubscript{M}:} Another way to support censored observations using MAE is by assigning a “best guess” value (the margin time) to each censored observation~\citep{haider_effective_2020}, using the KM estimator~\citep{kaplan_nonparametric_1958}. Given the event time is greater than the censoring time, the margin time is then a conditional expectation of the event time given some observation censored at $t_{i}$~\citep{qi2023effective}:
\begin{equation}
e_{\text{margin}}(t_i, D) = \mathbb{E}_t[e_i \,|\, e_i > t_i] = t_i + \frac{\int_{t_i}^{\infty} t \, S_{\text{KM}}(D)(t) \, dt}{S_{\text{KM}}(D)(t)}\text{,}
\end{equation}

\noindent where $S_{\text{KM}}(D)(t)$ in the KM estimator derived from the training data.~\cite{haider_effective_2020} proposed using a confidence weight,
$\omega_i = 1 - S_{\text{KM}}(D)(t_i)$, based on the margin value, which gives lower confidence for observations censored early and higher confidence for observations censored late ($\omega = 1$ for uncensored observations). Adopting this weighting scheme, the MAE-margin is then~\citep{qi2023effective}:
\begin{align}\label{eq:maemargin}
&\text{MAE-Margin}(\hat{t}_i, t_i, \delta_i)] \nonumber \\
&= \frac{1}{\sum_{i=1}^{N}\omega_i} \sum_{i=1}^{N} \omega_i \cdot | [(1 - \delta_i) \cdot e_{\text{margin}}(t_i) + \delta_i \cdot t_i] - \hat{t}_i |\text{.}
\end{align}

\textbf{MAE\textsubscript{PO}:} We can also substitute the unobserved event times with pseudo-observations~\citep{andersen_pseudo-observations_2010}. Let $\{t_i\}_{i=1}^{N}$ be i.i.d. draws of a random variable time, $T$, and let $\hat{\theta}$ be an unbiased estimator for the event time based on right-censored observations of $T$. The pseudo-observation for a censored subject is defined as:
\begin{equation}\label{eq:pseudo}
e_{\text{pseudo}}(t_i, D) = N \times \hat{\theta} - (N - 1) \times \hat{\theta}^{-i}\text{,}
\end{equation}

\noindent where $\hat{\theta}^{-i}$ is the estimator applied to the $N-1$ element dataset formed by removing that $i$-th observation. The pseudo-observation can be viewed as the contribution of observation $i$ to the unbiased event time estimation $\hat{\theta}$. Following~\cite{qi2023effective}, we estimate $\hat{\theta}$ using the KM estimator, \ie $\hat{\theta} = E_{t}[S_{\text{KM}}(D)(t)]$ and $\hat{\theta}^{-i} = E_{t}[S_{\text{KM}}(D^{-i})(t)]$ as unbiased estimators. After calculating the pseudo-observation values using Eq. \ref{eq:pseudo}, the MAE-pseudo-observation metric (MAE\textsubscript{PO}) can be calculated using the weighting 
scheme in Eq. \ref{eq:maemargin}.

\textbf{D-Calibration:} Distribution calibration assesses if the survival curve, $S(t)$, is calibrated~\citep{haider_effective_2020}, informing us to which extent we should trust its predicted probabilities. For each uncensored observation, $\bm{x}_{i}$, with some observed event time, $y_{i}$, we determine the probability $P(S(t\text{\textbar}\bm{x}{i}(t)))$ for that time based on $S(t)$. Following the example of~\cite{haider_effective_2020}, if $S(t)$ is D-Calibrated, we anticipate 10\% of the events to fall in the [90\%, 100\%] interval and another 10\% to fall in the [80\%, 90\%] interval. For all individual survival distributions, $d_{i}$, the set $\{S_{i}(d_{i})\}$ should exhibit a uniform distribution between zero and one, ensuring each 10\%-interval covers 10\% of $D$. We use a Pearson's $\chi^2$ test to check if the proportion of events in each interval is uniformly distributed within mutually exclusive and equal-sized intervals, as proposed by~\cite{haider_effective_2020}.

\textbf{C-Calibration}: Coverage calibration is a test to assess the alignment between the predicted credible intervals (CrIs) and the observed probability intervals~\citep{qi_using_2023}, \ie a p\% probability CrI should give individual CrIs with a probability of p\% to encompass an observations's likelihood of event. The coverage calibration procedure involves calculating observed coverage rates using CrIs with varying percentages, ranging from 10\% to 90\%. We follow the approach of~\cite{qi_using_2023} and use a Pearson's $\chi^2$ test to evaluate the calibration of the observed and expected coverage rates. The null hypothesis is that the observed coverages are consistent across all percentage groups. Models that show similarity between expected and observed event rates in subgroups are C-calibrated. This test is specific to models that can predict CrIs.

\section{Censoring algorithm}
\label{app:censoring_algorithm}

The algorithm \eqref{alg:cens_algo} is the pseudocode of our proposed censoring algorithm. The algorithm adds random and independent censoring of a specific percentage $C$ to the dataset $\mathcal{D}$.

\begin{algorithm}[!ht]
    \caption{Proposed censoring algorithm.}
    \label{alg:cens_algo}
    \begin{algorithmic}[1]
        \Require Uncensored dataset $\mathcal{D}$ and censoring percentage $C$.
        \State Sample $\mathcal{I}$ as a set of $C$ random indices from $\mathcal{D}$
        \State Let $e_i$ be the event time and $\delta_{i}$ the event indicator for $i$ in $\mathcal{D}$
        \ForAll{observation $i$ in $\mathcal{I}$}
        \State Generate a random integer \(c_i\) in the range $[1, e_i]$
        \State Set $t_i=c_i$ and $\delta_{i}=0$ in $\mathcal{D}$
        \EndFor
        \State \Return Censored dataset $\tilde{\mathcal{D}}$
    \end{algorithmic}
\end{algorithm}

\bibliographystyle{elsarticle-harv}\biboptions{authoryear}
\bibliography{references}

\end{document}